\documentclass[conference]{style/IEEEtran}

\usepackage[nolist]{acronym}
\usepackage{siunitx}
\usepackage{hyperref}
\usepackage{xspace}
\usepackage{lipsum}
\usepackage{graphicx}
\usepackage{enumitem}
\usepackage{algorithmic}
\usepackage[ruled,vlined]{algorithm2e}
\usepackage{color}
\usepackage{float}
\usepackage{mathtools}
\usepackage{subcaption}
\usepackage{amsmath}
\usepackage{amsfonts}
\usepackage{amsthm}
\usepackage{stmaryrd}
\usepackage{cleveref}
\usepackage{todonotes}
\usepackage{tikz}
\usetikzlibrary{arrows.meta}
\usetikzlibrary{matrix,arrows}
\usetikzlibrary{positioning}
\usetikzlibrary{decorations,decorations.text}
\usetikzlibrary{decorations.pathmorphing}
\usetikzlibrary{shapes,arrows,matrix, positioning,calc}
\usetikzlibrary{backgrounds}
\usetikzlibrary{fit}
\usepackage{xargs}
\usepackage{booktabs}
\usepackage{multirow}
\usepackage{misc/nicodemus}
\newcommand{\secpar}{\lambda}
\newcommand{\getsr}{\mathrel{\vbox{\offinterlineskip\ialign{
	\hfil##\hfil\cr
	\hspace{0.1em}$\scriptscriptstyle\$$\cr
	$\leftarrow$\cr
}}}} 
\makeatletter
\newcommand{\vast}{\bBigg@{3}}
\newcommand{\Vast}{\bBigg@{4}}
\makeatother
\newcommand{\bool}[1]{\llbracket #1\rrbracket}
\newcommand{\codecomment}[1]{{\footnotesize \textit{\# #1}}}

\crefname{assumption}{Assumption}{Assumptions}
\crefname{construction}{Construction}{Constructions}
\crefname{corollary}{Corollary}{Corollaries}
\crefname{conjecture}{Conjecture}{Conjectures}
\crefname{definition}{Definition}{Definitions}
\crefname{exmaple}{Example}{Examples}
\crefname{figure}{Figure}{Figures}
\crefname{lemma}{Lemma}{Lemmata}
\crefname{observation}{Observation}{Observations}
\crefname{proposition}{Proposition}{Propositions}
\crefname{remark}{Remark}{Remarks}
\crefname{section}{Section}{Sections}
\crefname{theorem}{Theorem}{Theorems}
\crefname{table}{Table}{Tables}
\crefname{appendix}{Appendix}{Appendices}

\newcommand{\pcfor}{\textbf{for }}

\newcommandx{\pcforin}[2]{\textbf{for }{#1}\textbf{ in }{#2}\colon}
\newcommand{\pcif}{\textbf{if }}
\newcommand{\pcor}{\textbf{or }}
\newcommand{\pcand}{\textbf{and }}

\newcommand{\pcnot}{\textbf{not }}

\newcommand{\pcreturn}{\textbf{return }}

\newcommand{\true}{\textbf{true}}
\newcommand{\false}{\textbf{false}}

\newcommand{\pcabort}{\textbf{abort}}

\newcommand{\idx}{idx}

\newcommand{\cntUser}{u}
\newcommand{\users}{\mathcal{U}}
\newcommand{\userdata}{\widehat{\data}_{\cntUser}}
\newcommand{\server}{S}

\newcommand{\treeroot}{\Psi}

\newcommand{\pp}{\mathsf{pp}}

\newcommand{\SNARK}{\Pi}
\newcommand{\SNARKSetup}{\SNARK.\mathsf{Setup}}
\newcommand{\SNARKProve}{\SNARK.\mathsf{Prove}}
\newcommand{\SNARKVrfy}{\SNARK.\mathsf{Vrfy}}

\newcommand{\crs}{\mathsf{pp}}
\newcommandx{\train}[1][1=:~~]{\textit{train#1}}
\newcommandx{\unlearn}[1][1=:~~]{\textit{unlearn#1}}
\newcommandx{\mode}[2][1=:~~,2=]{\textit{mode$_{#2}$#1}}

\newcommand{\sig}{\sigma}
\newcommandx{\sigA}[1][1=]{\sig_{#1}}
\newcommandx{\sigS}[1][1=]{\sig_{S#1}}

\newcommandx{\model}[1][1=]{m_{#1}}

\newcommandx{\data}[1][1=]{D_{#1}}
\newcommandx{\alldata}{\mathcal{D}}
\newcommandx{\listDatasets}{\mathcal{L}_{\data}}
\newcommandx{\listHashUnlearned}{\mathcal{L}_{h}}
\newcommandx{\dataUnlearn}[1][1=]{U_{#1}}
\newcommandx{\dataUnlearnP}[1][1=]{{U_{#1}'}^{+}}
\newcommandx{\modelD}[1][1=]{\model_{\data[#1]}}

\newcommandx{\proofModel}[1][1=]{\ensuremath{\rho}_{#1}}
\newcommandx{\proofM}[1][1=]{\ensuremath{\pi}_{\model[#1]}}
\newcommandx{\proofD}[1][1=]{\ensuremath{\pi}_{\data[#1]}}
\newcommandx{\proofI}[1][1=]{\ensuremath{\pi}_{0}}
\newcommandx{\proofT}[1][1=]{\ensuremath{\pi}_{#1}}
\newcommandx{\proofU}[1][1=]{\ensuremath{\pi}_{#1}}

\newcommand{\witness}{\omega}

\newcommandx{\listCom}{\mathcal{L}_{\com}}
\newcommandx{\listModel}{\mathcal{L}_{\model}}
\newcommandx{\listUp}{\mathcal{L}_{\up}}

\newcommand{\relation}{R}
\newcommand{\statement}{\phi}

\newcommandx{\ledgerEntry}[1][1=]{L_{#1}}
\newcommand{\append}{\mathsf{append}}

\newcommand{\idxf}{\mathsf{index}}

\newcommand{\Init}{\mathsf{Init}}
\newcommand{\VrfyInit}{\mathsf{VerifyInit}}
\newcommand{\ProveUnlearn}{\mathsf{ProveUnlearning}}
\newcommand{\ProvePrivUnlearn}{\mathsf{ProveNonMembership}}
\newcommand{\VrfyUnlearn}{\mathsf{VerifyUnlearn}}
\newcommand{\VrfyPubUnlearn}{\mathsf{VerifyUnlearning}}
\newcommand{\VrfyPrivUnlearn}{\mathsf{VerifyNonMembership}}
\newcommand{\ProveUpdate}{\mathsf{ProveTraining}}
\newcommand{\VrfyUpdate}{\mathsf{VerifyTraining}}

\newcommand{\Commit}{\mathsf{Commit}}

\newcommandx{\up}[1][1=]{\mathsf{up}_{#1}}

\newcommandx{\hashsData}[1][1=]{\mathcal{H}_{\data[#1]}}
\newcommandx{\hashsDataP}[1][1=]{\mathcal{H}_{\data[#1]}'}
\newcommandx{\hashsDataAdd}[1][1=]{\mathcal{H}_{\dataAdd[#1]}}
\newcommandx{\hashsDataAddP}[1][1=]{\mathcal{H}_{\dataAdd[#1]}'}
\newcommandx{\hashsUnlearned}[1][1=]{\mathcal{H}_{\dataUnlearn[#1]}}
\newcommandx{\hashsUnlearnedP}[1][1=]{\mathcal{H}_{\dataUnlearn[#1]}'}
\newcommandx{\hashsUnlearnedAdd}[1][1=]{\mathcal{H}_{\dataUnlearnAdd[#1]}}
\newcommandx{\hashsUnlearnedAddP}[1][1=]{\mathcal{H}_{\dataUnlearnAdd[#1]}'}
\newcommandx{\hashsAllData}[1][1=]{\mathcal{H}_{\data[all]}}

\newcommandx{\dataAdd}[1][1=]{D^+_{#1}}
\newcommandx{\dataUnlearnAdd}[1][1=]{U^+_{#1}}
\newcommandx{\dataAddP}[1][1=]{D_{#1}^+{}'}
\newcommandx{\dataUnlearnAddP}[1][1=]{U_{#1}^+{}'}

\newcommandx{\stS}[1][1=]{\mathsf{st}_{S,#1}}
\newcommandx{\stA}[1][1=]{\mathsf{st}_{A,#1}}
\newcommandx{\stU}[1][1=]{\mathsf{st}_{u,#1}}

\newcommand{\extractor}{\mathcal{E}}
\newcommand{\aux}{\mathsf{aux}}

\newcommand{\relationI}{R_{I}}
\newcommand{\relationT}{R_{T}}
\newcommand{\relationU}{R_{U}}

\newcommand{\ppI}{\pp_{I}}
\newcommand{\ppT}{\pp_{T}}
\newcommand{\ppU}{\pp_{U}}

\newcommand{\circuitInit}{C_{I}}
\newcommand{\circuitTraining}{C_{T}}
\newcommand{\circuitUnlearning}{C_{U}}

\newcommand{\x}{x}

\newcommand{\y}{y}

\newcommand{\datasample}{d}
\newcommand{\ComputeTreePath}{\mathsf{ComputeChainPath}}
\newcommand{\VrfyTreePath}{\mathsf{VerifyChainPath}}

\newcommandx{\dUnlearn}[1][1=]{\datasample_{#1}}
\newcommandx{\proofUnlearn}[1][1=]{\ensuremath{\pi}_{\cntUser,\dUnlearn[#1]}}

\newcommand{\Hash}{\mathsf{Hash}}

\newcommand{\lenHash}{\kappa}
\newcommandx{\hashData}[1][1=]{h_{\data[#1]}}

\newcommandx{\hashDataUnlearn}[1][1=]{h_{\dataUnlearn[#1]}}
\newcommandx{\hashDataSample}[1][1=]{h_{\datasample_{#1}}}

\newcommandx{\hashModel}[1][1=]{h_{\modelD[#1]}}
\newcommandx{\hashLedger}[1][1=]{h_{\ledgerEntry[#1]}}

\newcommand{\HashData}{\mathsf{HashData}}
\newcommand{\AppendHashData}{\mathsf{AppendHashData}}
\newcommand{\AppendHashDataUnlearn}{\mathsf{AppendHashData}}

\newcommandx{\hData}[1][1=]{h_{\data[#1]}}
\newcommandx{\hModel}[1][1=]{h_{\model[#1]}}
\newcommandx{\hUnlearn}[1][1=]{h_{\dataUnlearn[#1]}}
\newcommandx{\hDataP}[1][1=]{h'_{\data[#1]}}
\newcommandx{\hModelP}[1][1=]{h'_{\model[#1]}}
\newcommandx{\hUnlearnP}[1][1=]{h'_{\dataUnlearn[#1]}}
\newcommandx{\hStF}[1][1=]{h_{\stF[#1]}}
\newcommand{\dataempty}{\datasample_\emptyset}

\newcommand{\HashDataSample}{\mathsf{HashDataRecord}}
\newcommand{\HashDataUnlearn}{\mathsf{HashData}}
\newcommand{\HashModel}{\mathsf{HashModel}}
\newcommand{\HashState}{\mathsf{HashState}}

\newcommand{\hashs}{\mathcal{H}}

\newcommand{\advA}{\mathcal{A}}

\newcommand{\negl}{\mathsf{negl}}

\newcommand{\SecurityUnlearn}{\mathsf{GameUnlearn}}
\newcommandx{\game}[1][1=]{\mathsf{G}_{#1}}

\newcommand{\admissibleFunction}{f}
\newcommand{\admissibleFunctionsSet}{\mathcal{F}}
\newcommand{\unlearningFunction}{\admissibleFunction_{U}}
\newcommand{\trainingFunction}{\admissibleFunction_{T}}
\newcommand{\initFunction}{\admissibleFunction_{I}}
\newcommand{\protocolF}{\Phi_\admissibleFunction}
\newcommand{\seed}{\mathsf{pp}_f}
\newcommandx{\stF}[1][1=]{\mathsf{st}_{\admissibleFunction#1}}

\newtheorem{theorem}{Theorem}

\newtheorem{definition}{Definition}

\newcommand*{\ie}{{\em i.e.,}\xspace}
\newcommand*{\etal}{{\em et al.}\@\xspace}
\newcommand*{\eg}{{\em e.g.,}\@\xspace}

\newcommand{\epochs}{E}

\newcommand{\cntepoch}{e}

\newcommand{\weightsmodel}{\theta}
\newcommand{\weightsInitial}{\theta_{initial}}
\newcommand{\parupdate}{\ensuremath{\Delta}}

\newcommandx{\com}[1][1=]{\mathsf{com}_{#1}}
\newcommand{\Setup}{\mathsf{Setup}}

\newcommand{\pub}{\mathsf{pub}}

\newcommandx{\comNonMember}[1][1=]{\ensuremath{\Psi}^{\scalebox{0.5}[1.0]{\( - \)}}_{#1}}

\newcommandx{\cert}[1][1=]{c_{#1}}

\newcommand{\cntIteration}{i}

\newcommand{\cntAdd}{k}
\newcommand{\cntDelete}{j}
\newcommandx{\noDeleteQueriesIteration}[1][1=\cntIteration]{t_{#1}}
\newcommand{\noIterations}{\ell}

\newcounter{ctr}

\newcommand{\mypar}[1]{\medskip\noindent\textbf{#1.}\xspace}
\newcommand{\mysep}{2pt}

\hypersetup{
    colorlinks = false,
    hidelinks = true
}

\usepackage[most]{tcolorbox}
\tcbset{%
  enhanced,
  colback=black!2,
  colframe=black!60,
  sharp corners,
  boxrule=0pt,
  frame hidden,
  borderline west={1.5mm}{0mm}{black!15}
}
\begin{acronym}
    \acro{MI}{\emph{Model Inversion}}
    \acro{MIA}{\emph{Membership Inference Attack}}
    \acro{SISA}{\emph{Sharded, Isolated, Sliced, and Aggregated}}
    \acro{R1CS}{\emph{Rank-1 Constraint Systems}}
    \acro{MPC}{\emph{Multi-Party Computation}}
    \acro{HE}{\emph{Homomorphic Encryption}}
    \acro{ADSs}{\emph{Authenticated Data Structures}}
    \acro{VC}{\emph{Verified Computation}}
\end{acronym}

\begin{document}

\title{Verifiable and Provably Secure Machine Unlearning}

\author{%
Thorsten Eisenhofer\IEEEauthorrefmark{1}, 
Doreen Riepel\IEEEauthorrefmark{2}, 
Varun Chandrasekaran\IEEEauthorrefmark{3}, \\
Esha Ghosh\IEEEauthorrefmark{4}, 
Olga Ohrimenko\IEEEauthorrefmark{5}, 
Nicolas Papernot\IEEEauthorrefmark{6}\vspace*{0.15cm} \\ 
BIFOLD \& TU Berlin\IEEEauthorrefmark{1}, 
CISPA Helmholtz Center for Information Security\IEEEauthorrefmark{2},\\
University of Illinois Urbana-Champaign\IEEEauthorrefmark{3},
Microsoft Research\IEEEauthorrefmark{4},\\
The University of Melbourne\IEEEauthorrefmark{5},
University of Toronto \& Vector Institute\IEEEauthorrefmark{6} 
}

\IEEEoverridecommandlockouts
\IEEEpubid{\makebox[\columnwidth]{\hfill} \hspace{\columnsep}\makebox[\columnwidth]{ }}

\maketitle

\IEEEpubidadjcol

\begin{abstract}
Machine unlearning aims to remove points from the training dataset of a machine learning model after training: e.g., when a user requests their data to be deleted. While many unlearning methods have been proposed, none of them enable users to audit the procedure. Furthermore, recent work shows a user is unable to verify whether their data was unlearnt from an inspection of the model parameter alone. Rather than reasoning about parameters, we propose to view {\em verifiable unlearning} as a security problem. To this end, we present the first cryptographic definition of verifiable unlearning to formally capture the guarantees of an unlearning system. In this framework, the server first computes a proof that the model was trained on a dataset~$D$. Given a user's data point~$d$ requested to be deleted, the server updates the model using an unlearning algorithm. It then provides a proof of the correct execution of unlearning {\em and} that $d \notin D'$, where $D'$ is the new training dataset (i.e., $d$ has been removed). Our framework is generally applicable to different unlearning techniques that we abstract as \emph{admissible functions}. We instantiate a protocol in the framework, based on cryptographic assumptions, using SNARKs and hash chains. Finally, we implement the protocol for three different unlearning techniques and validate its feasibility for linear regression, logistic regression, and neural networks.
\end{abstract}

\section{Introduction}
\label{sec:intro}

The right to be forgotten entitles individuals to self-determine the possession of their private data and compel its deletion. In practice, this is now mandated by regulations like the GDPR~\cite{misc-gdpr}, CCPA~\cite{misc-ccpa}, or PIPEDA~\cite{misc-pipeda}. Consider the case where a company or service provider collects data from its users. These regulations allow users to request a deletion of their data and legally compels the company to fulfil the request. However, this is challenging when the data is used for downstream analyses, \eg training machine learning (ML) models, where the relationship between model parameters and the data used to obtain them is complex~\cite{koh-17-understanding}. In particular, ML models are known to memorize information from their training set~\cite{feldman-20-learning,brown-21-memorization}, resulting in attacks on the privacy of training data~\cite{carlini-19-secret,leino-20-stolen}.

Thus, techniques have been introduced for \textit{unlearning}: a trained model is updated to remove the influence a training point had on the model's parameters and predictions~\cite{cao-15-towards}. Regardless of the particular approach, existing techniques~\cite{wu-18-dizk,bourtoule-19-machine,guo-19-certified,golatkar-20-eternal,graves-21-amnesiac,baumhauer-20-machine,sekhari-21-remember} suffer from one critical limitation: they are unable to {\em provide the user with a proof that their data was indeed unlearnt.} 
This is problematic because dishonest service providers may falsify unlearning to avoid paying the large computational costs or to maintain model utility~\cite{sener-17-active,ghorbani-19-data}.
 
Additionally, verifying that a point is unlearnt is non-trivial \emph{from the user's perspective}. A primary reason is that users (or third-party auditors) cannot determine whether a data point is unlearnt (or not) by comparing the model's predictions or parameters before and after the claimed unlearning. The complex relationship between training data, models' parameters, and their predictions make it difficult to isolate the effects of any training point. In fact, prior work~\cite{shumailov-21-manipulating,thudi-22-necessity} demonstrates that a model's parameters can be identical when trained with or without a data point. To address these concerns, we propose {\em a cryptographic approach to verify unlearning}. Rather than trying to verify unlearning by examining changes in the model, we ask the service provider (\ie the server) to present a cryptographic proof that an {\em agreed-upon} unlearning process was executed. This leads us to view unlearning as a security problem that we aim to solve with formal guarantees.

In this paper, we propose {\em the first formal framework} of verifiable machine unlearning. The framework describes the interface of an unlearning protocol in an algorithmic manner and also defines a game-based security notion which allows to prove security of protocol instantiations based on cryptographic assumptions.
In order to capture the desired security goals, we find that the definition needs to ensure consistency of data during training and unlearning, {\em and} across model updates and evolving datasets as it requires a user to be able to verify that their data was not re-added at later stages.
Therefore, we formalize unlearning in our framework as an iteration-based protocol; this requires the server to prove that it has honestly updated the model and dataset in each iteration, either due to training with new data or unlearning previously used data. Only then does the user have sufficient guarantees about deletion of their data. Under this definition, we can instantiate protocols using any unlearning technique and any cryptographic primitives that have appropriate security guarantees.
We capture the relationship between models and datasets via initialization, training, and unlearning functions with an abstraction that we call \emph{admissible functions}.

In our framework, we identify the following guarantees that need to be satisfied: (a) the model was trained \emph{conceptually} from some dataset (more details in \S~\ref{sec:framework-overview}), and (b) the user's data point is not present in this dataset. 
Thus, the framework has two major components. First, the server computes a {\em proof of training} whenever data points are added to the model's training data: this establishes that it trained the model on a particular dataset. 
Second, when a user submits a request to unlearn a specific data point, the server computes a {\em proof of unlearning} that proves that the model and its conceptual training set was updated addressing the request. Additionally, the server provides the user with a \emph{proof} that their data point is not part of the updated training set. By linking the executions of those components across iterations, these proofs also ensure that no data point can be added back to the training set \emph{after} it was unlearnt.

We present a fully instantiated protocol in our framework. This instantiation uses SNARK-based verifiable computation as a generic approach for the proof of model updates induced by training or unlearning, and hash chains for the proof of non-membership in a model's training set. Under our security definition, we formally prove the correctness and security of this instantiation generically for any training and unlearning algorithms covered by the abstraction of admissible functions.

Finally, we provide the first implementation of verifiable unlearning based on cryptographic primitives.
In particular, we instantiate the SNARK with the Spartan~\cite{setty-20-spartan} proof system for verifiable training and unlearning. 
We consider three unlearning techniques: retraining-based unlearning, amnesiac unlearning~\cite{graves-21-amnesiac} and optimization-based unlearning~\cite{jang-22-knowledge, warnecke-21-unlearning}.
We demonstrate the versatility and scalability of our construction on a variety of binary classification tasks from the PMLB benchmark suite~\cite{romano-21-pmlb}, using linear regression, logistic regression, and simple neural networks.

\mypar{Contributions} We make the following contributions:
\begin{itemize}[topsep=2pt, itemsep=\mysep, partopsep=\mysep, parsep=\mysep]
\item \emph{Formal framework.} We introduce a framework to construct protocols for verifiable machine unlearning. Our framework is designed to be general enough to capture different unlearning algorithms and secure primitives. It models verifiable unlearning as a 2-party protocol (executed between a service provider and its users).

\item \emph{Security definition of verifiable machine unlearning.} We then propose a formal security definition of a verifiable machine unlearning scheme. This game-based definition allows to prove the security of unlearning protocols within our framework. 

\item \emph{Instantiation.} We present a fully instantiated protocol. This construction is based on a generic interface of admissible functions for training and unlearning and thus applicable to any training and unlearning algorithm (as captured by the abstraction).

\item \emph{Practical implementation.} 
Finally, we provide the first implementation of verifiable unlearning based on cryptographic primitives.
We study its applicability to three unlearning techniques, different classes of ML models, and benchmark datasets.

\end{itemize}
\section{Background}
\label{sec:background}
In this section, we discuss the preliminaries needed to understand the contributions of our work.

\mypar{Notation} Throughout the paper, let $\lambda$ denote the security parameter. We call a function negligible in $\secpar$---denoted by $\negl(\secpar)$---if it is smaller than the inverse of any polynomial for all large enough values of $\secpar$. $[m:n]$ denotes the set $\{m,m + 1, ..., n\}$ for integers $m < n$. For $m = 1$, we simply write $[n]$. $y \gets \mathsf{M}(x_1, x_2,\ldots)$ denotes that on input $x_1, x_2, \ldots$, the probabilistic algorithm $\mathsf{M}$ returns $y$.
An adversary $\advA$ is a probabilistic algorithm, and is \textit{efficient} or Probabilistic Polynomial-Time (PPT) if its run-time is bounded by some polynomial in the length of its input. We will use code-based games, where $\Pr[\mathsf{G}\Rightarrow1]$ denotes the probability that
the final output of game $\mathsf{G}$ is 1.

\subsection{Machine Learning Preliminaries}
\label{sec:background-ml}

We start with the required background on machine learning and introduce several techniques for unlearning.

\mypar{Supervised Machine Learning} Supervised machine learning (ML) is the process of learning a parameterized function $f_{\theta}$ (often called a \emph{model}) that is able to predict an output (from the space of outputs $\mathcal{Y}$) given an input (from the space of inputs $\mathcal{X}$), \ie $f_\theta: \mathcal{X} \rightarrow \mathcal{Y}$. Commonly learnt functions include linear regression, logistic regression, and neural networks.

The parameters of this function are typically optimized using methods such as stochastic gradient descent (SGD).
Let $\weightsInitial$ be randomly initialized parameters and $\data = \{d_1,\ldots,d_n\}$ a set of training data points, where each $\datasample = (\datasample_x, \datasample_y) \in \mathcal{X}\times\mathcal{Y}$.
During training we iteratively update parameters as $\weightsmodel' \coloneqq \weightsmodel - \eta \nabla_{\weightsmodel} \mathcal{L}(f_{\weightsmodel}(d_x), d_y)$ for points $\datasample \in \data$ where $\mathcal{L}$ is a suitably chosen loss function (\eg cross-entropy loss) and $\eta$ the \emph{learning rate}.

In SGD, the update is calculated for a randomly chosen input $\datasample \in \data$ in each step. In practice, this is often extended to \emph{batches} of data points in order to reduce the variance of each update.
Often, multiple passes (called \emph{epochs}) are repeated through the dataset. 
We can describe the full process of \emph{training} a model $\model$ (interchangeably used with $\weightsmodel_{\model}$) by
\vspace{-0.15em}
$$\weightsmodel_{\model}\coloneqq \weightsInitial+\sum_{\cntepoch\in[\epochs]}\sum_{\datasample \in \data }\parupdate_{\cntepoch,\datasample}\, ,$$
with $\epochs$ being the number of epochs and $\parupdate_{\cntepoch,\datasample}$ the update on the model's parameter from data point $\datasample$ in epoch $\cntepoch$.

\mypar{Machine Unlearning} In machine unlearning the goal is to design algorithms that enable an ML model (specifically, its parameters) to forget the contribution of a (subset of) data point(s).
The canonical approach for this is to naively retrain the model from scratch. Hence, removing a data point $d^*$  from a model $\model$ with \emph{retraining-based} unlearning can be described as
\vspace{-0.15em}
$$\weightsmodel_{\model'}\coloneqq \weightsInitial+\sum_{\cntepoch\in[\epochs]}\sum_{\datasample\in\data\setminus{\{\datasample^*\}}}\parupdate_{\cntepoch,\datasample} \, .$$
As the resulting model $\weightsmodel_{\model'}$ is completely devoid of data point $d^*$ (by construction), this is an example for {\em exact unlearning}~\cite{bourtoule-19-machine,cao-15-towards,wu-20-deltagrad,neel-21-descent}, which is desirable but often prohibitively expensive.

More practical unlearning techniques where the contribution of a data point cannot be \emph{completely} removed and the guarantees tolerate some error~\cite{guo-19-certified,chen-21-graph,thudi-22-unrolling,baumhauer-20-machine,sekhari-21-remember,golatkar-20-eternal,graves-21-amnesiac} are commonly referred to as  {\em approximate unlearning}.
An example for this is \emph{amnesiac unlearning}~\cite{graves-21-amnesiac}. Given a model $\model$, removing a data point $d^*$ with amnesiac unlearning means that we compute
\vspace{-0.15em}
$$\weightsmodel_{\model'}\coloneqq \weightsmodel_{\model} - \sum_{\cntepoch \in [\epochs]}\parupdate_{\cntepoch, \datasample^*} \, .$$ 
In other words, to unlearn data point $d^*$, we remove the updates to the model's parameters that were \emph{directly} computed on that data point for all training epochs $\epochs$.
Yet, amnesiac unlearning only provides approximate guarantees since updates from unlearnt data points indirectly influence updates from later points during the iterative nature of the training~\cite{thudi-22-unrolling}.

Other approaches for approximate unlearning formulate unlearning as an optimization problem~\cite{jang-22-knowledge, warnecke-21-unlearning} (similar to training). In every step, instead of reducing the loss of a data point, we increase it (\ie \emph{unlearn} this data point). We refer to this as \emph{optimization-based} unlearning. Formally, we iteratively compute an update $\parupdate_{\cntepoch,\datasample^*}$ for the current model that is \emph{subtracted} from its parameters:
\vspace{-0.15em}
$$\weightsmodel_{\model'}\coloneqq \weightsmodel_{\model} - \sum_{\cntepoch \in [\hat{\epochs}]}\parupdate_{\cntepoch, \datasample^*}\, ,$$
where $\hat{\epochs}$ denotes the number of \emph{unlearning epochs} and $\parupdate_{\cntepoch, \datasample^*}$ the update from data point $\datasample^*$ in epoch $\cntepoch$. For model parameters $\weightsmodel$ (in epoch $e$), we define $\parupdate_{\cntepoch, \datasample^*} \coloneqq -\hat{\eta} \nabla_{\weightsmodel} \mathcal{L} (f_{\weightsmodel}(d^*_x), d^*_y)$ with \emph{unlearning rate} $\hat{\eta}$ and loss function $\mathcal{L}$.

\subsection{Cryptographic Preliminaries}
\label{subsec:crypto_primitives}

We want to provide the user with a cryptographic proof that their data were indeed deleted. Therefore, we require several cryptographic primitives that we introduce next.

\mypar{Collision-Resistant Hash Functions} A family of hash function $\mathcal{H}:\{0,1\}^\lambda \times\{0,1\}^n \rightarrow \{0,1\}^\lenHash$ is collision-resistant if 
it is length-compressing, \ie $\lenHash < n$ and it is hard to find collisions, \ie for all PPT 
adversaries $\advA$, and for all security parameters $\secpar$,
\begin{center}
\scalebox{0.9}{
$\Pr\Bigg[\!\!\!\begin{array}{c} k\getsr \{0,1\}^\lambda ; (x_0, x_1) \leftarrow \advA(1^\secpar,\mathcal{H}_k):  \\ x_0 \neq x_1 \wedge \mathcal{H}_k(x_0) = \mathcal{H}_k(x_1) \end{array}\Bigg]\leq \negl(\lambda) \, .$
}
\end{center}
In this work, we will denote $\Hash$ as a randomly chosen function from $\mathcal{H}$.

\mypar{Proof Systems} An interactive proof system describes a protocol between a \emph{prover} and a \emph{verifier}, where the prover wants to convince the verifier that some statement $\statement$ for a given polynomial time decidable relation $\relation$ is true. The prover holds a witness $\witness$ for the statement. 
In this work we are concerned with \emph{non-interactive} proof systems. A Succinct Non-Interactive Argument of Knowledge (SNARK) allows the prover to non-interactively prove the statement with a short (or succinct) cryptographic proof which can be verified in time sublinear in the size of the statement. We denote a SNARK by $\SNARK$ and define it by the three algorithms $(\SNARKSetup,\SNARKProve,\SNARKVrfy)$:
\begin{description}[topsep=5pt, itemsep=2pt, partopsep=2pt, parsep=2pt]
\item $\crs\gets\SNARKSetup(1^\secpar,R)$: The setup algorithm outputs public parameters $\pp$ for a polynomial-time decidable relation $R$.
\item $\pi\gets\SNARKProve(R,\crs,\phi,\witness)$: The prover algorithm takes as input $\crs$ and $(\phi,\witness)\in R$ and returns an argument $\pi$, where $\phi$ is termed the statement and $\witness$ the witness. 

\item $b\gets\SNARKVrfy(R,\crs,\phi,\pi)$: The verification algorithm takes $\crs$, a statement $\phi$ and an argument $\pi$ and returns a bit $b$, where $b=1$ indicates success and $b=0$ indicates failure.
\end{description}

\smallskip
\noindent \textit{Perfect Completeness.}  Given any true statement, an honest prover should be able to convince an honest verifier. More formally, let $\mathcal{R}$ be a sequence of families of efficiently decidable relations $\relation$. For all $R\in\mathcal{R}$ and $(\phi,\witness)\in R$
\vspace*{3mm}

{\centering
\hspace*{0.35cm}\scalebox{0.9}{
$\Pr\Bigg[ \SNARKVrfy(R,\crs,\phi,\pi)\Bigg|\!\! \begin{array}{l} \crs\gets\SNARKSetup(1^\secpar,R); \\ \pi\gets\SNARKProve(R,\crs,\phi,\witness) \end{array}\!\!\!\Bigg]=1 \, .$}}
\vspace*{3mm}

\label{def:snark-completeness}
\medskip
\noindent \textit{Computational Soundness.} 
We say that $\SNARK$ is sound if it is not possible to prove a false statement. Let $L_R$ be the language consisting of statements for which there exists corresponding witnesses in $R$. For a relation $R\sim\mathcal{R}$, we require that for all non-uniform PPT adversaries $\advA$
\vspace*{3mm}

{\centering
\hspace*{0.35cm}\scalebox{0.9}{
$\Pr\Bigg[\!\!\!\begin{array}{c} \phi \notin L_R ~\pcand \\ \SNARKVrfy(R,\crs,\phi,\pi)\end{array}\!\!\Bigg|\!\! \begin{array}{l} \crs\gets\SNARKSetup(1^\secpar,R); \\ (\phi,\pi)\gets\advA(R,\crs) \end{array}\!\!\!\Bigg]\leq \negl(\lambda) \, .$
}}
\vspace*{3mm}
\label{def:snark-soundness}

We further define the notion of witness extractability or knowledge soundness.

\medskip
\noindent \textit{Computational Knowledge Soundness.} $\SNARK$ satisfies computational knowledge soundness if there exists an extractor that can compute a witness whenever the adversary produces a valid argument. Formally, for a relation $R\sim\mathcal{R}$, we require that for all non-uniform PPT adversaries $\advA$ there exists a non-uniform PPT extractor $\extractor$ such that
\vspace*{3mm}

{\centering
\hspace*{0.25cm}\scalebox{0.84}{
$\Pr\Bigg[\!\!\begin{array}{c} (\phi,\witness) \notin R ~\pcand \\ \SNARKVrfy(R,\crs,\phi,\pi)\end{array}\!\!\Bigg|\!\! \begin{array}{l} \crs\gets\SNARKSetup(1^\secpar,R); \\ ((\phi,\pi);\witness)\gets(\advA\|\extractor)(R,\crs) \end{array}\!\!\Bigg]\leq \negl(\lambda) \, .$
}}
\vspace*{3mm}

A SNARK $\SNARK$ is secure if it satisfies perfect completeness, computational soundness and knowledge soundness.
\section{Verifiable Machine Unlearning}
\label{sec:overview}

We consider the following ecosystem: there are many \emph{users}~$\users$, each of whom has access to a set of data points (or dataset).
They share their data with a \emph{server}~$\server$ which uses it to learn an ML model. Users can send requests to either delete or add new data. 
We assume the server is malicious and may not faithfully execute unlearning requests. This may be because the server is unwilling to tolerate model performance degradation after data deletion~\cite{sener-17-active,ghorbani-19-data}, or pay the computational penalty associated with model updating~\cite{bourtoule-19-machine, graves-21-amnesiac}.
Our goal is to develop a method to verify whether the server is adhering to users’ requests.

Before presenting our framework, we will first review conceptually simpler constructions. Specifically, we will discuss three naive approaches {\bf A1-A3}, and explain why they are insufficient. From this, we will derive necessary criteria {\bf D1-D3} that form the basis for our framework.

\mypar{{\bf A1}: Proof via Model Parameters} As a first step towards verifiable unlearning, assume the server holds a dataset on which it trained an ML model. To prove that it has unlearnt a specific data point, the server may provide the user with the trained model parameters (including random seeds) and the entire dataset. The user could then locally retrain the model and compare the resulting parameters with the server’s, or apply influence techniques \cite{koh-17-understanding} to assess whether their data point contributes to the model’s parameters. However, both of these methods suffer from a fundamental problem: it is possible to arrive at the same model parameters even if data was deleted. For example, recent work by Shumailov et al. \cite{shumailov-21-manipulating} and Thudi et al. \cite{thudi-22-necessity} describe how a user's contribution (towards model parameters) can be approximated from other entries in a dataset, rendering such approaches insufficient: the server can claim to have obtained the exact same model parameters from a number of different datasets.

\mypar{{\bf A2}: One-Shot Verified Unlearning} 
To account for this, we could extend {\bf A1} as follows: Instead of providing the user the model parameters and dataset only \emph{after} their data is (claimed to be) unlearnt, we require the server to \emph{prove} that it has executed a pre-specified unlearning algorithm. At this point, we do not want to delve into the specifics of what such a proof would entail, but one could envision verified computation techniques such as cryptographic methods or trusted-execution environments.
Instead, we want to point out some assumptions that are implicitly made; namely, we have not specified how the initial model (to which the unlearning algorithm is applied) was trained. For instance, we might assume that the model was truthfully trained which is commonly assumed when evaluating the efficacy of unlearning algorithms~\cite{bourtoule-19-machine, graves-21-amnesiac, warnecke-21-unlearning}. However, since we are considering a scenario where the server may behave maliciously during the unlearning process, we can not automatically assume that the training was conducted honestly either.

\mypar{{\bf A3}: A Naive Iteration-based Protocol} Therefore, we also need to require the server not only to prove the execution of the unlearning algorithm but also the training process itself. As a result, we aim to capture the \emph{evolution} of an ML model across multiple iterations of training and unlearning. For instance, after one user asked to unlearn their data point, another user might share new data with the server. However, this approach remains insufficient: a server could prove in one iteration that it has deleted the first user’s data, but then, in the next training iteration, it might simply reintroduce the same data point. This needs to be captured and a protocol needs additional mechanisms to avoid such behavior.

\mypar{Desiderata}
From the discussion thus far, we unearth three main requirements to achieve our goal of verifiable unlearning.
\begin{itemize}[topsep=2pt, itemsep=\mysep, partopsep=\mysep, parsep=\mysep]
\item[{\bf D1.}] The user must be able to verify that the unlearning algorithm was correctly executed and that the requested data point has been removed.
\item[{\bf D2.}] The user must be able to verify the model’s evolution, including the execution of training algorithms.
\item[{\bf D3.}] The system must ensure that the server cannot reintroduce previously unlearned data (unless explicitly requested to do so).
\end{itemize}

\section{Our Framework}\label{sec:framework-overview}
\begin{figure*}[t]
\centering
\hspace*{-0.15cm}
\resizebox{18.5cm}{!}{
	\begin{tikzpicture}
	\matrix (m0) [matrix of math nodes,row sep=-0.3em,column sep=0em,minimum width=0.1em,minimum height=1.2em,ampersand replacement=\&,
	column 1/.style={anchor=base west},
	column 4/.style={anchor=base west},
	column 7/.style={anchor=base west}] {
~~\textbf{Users}~\users~ \{\pub,\,\userdata\sim\alldata\}_{\cntUser\in\users} \& \hspace*{1.0cm} \& \hspace*{2.1cm} \& ~~~~\textbf{Server}~ \server~(\pub) \;  \hspace*{0.4cm}
\\
\; \& \; \& \; \& \; \& \; 
\\
\; \& \; \& \; \& \; \& \; 
\\
\; \& \; \& \; \& \; \& \; 
\\
\pcif \pcnot \VrfyInit(\pub,\com[0],\proofModel[0])\colon \& \; \& \; \& (\stS[0],\model[0],\com[0],\proofModel[0])\gets\Init(\pub) \& \; 
\\
\quad \pcabort \& \; \& \; \& \dataAdd[0]\coloneqq\emptyset,~ \dataUnlearnAdd[0]\coloneqq\emptyset  \& \; 
\\
\; \& \; \& \; \& \; \& \; 
\\
\; \& \; \& \; \& \dataAdd[\cntIteration]\coloneqq\dataAdd[\cntIteration-1],~ \dataUnlearnAdd[\cntIteration]\coloneqq\dataUnlearnAdd[\cntIteration-1] \& \; 
\\
\hspace*{0.8cm}\codecomment{add data points} \& \; \& \; \& \; \& \; 
\\
\hspace*{0.9cm}\cntAdd\text{-th query} \& \; \& \; \& \dataAdd[\cntIteration]\coloneqq\dataAdd[\cntIteration]\cup\{(\cntUser,d_{\cntIteration,\cntAdd})\} \& \; 
\\
\; \& \; \& \; \& \; \& \; 
\\
\hspace*{0.8cm}\codecomment{remove data points} \& \; \& \; \& \; \& \; 
\\
\hspace*{0.9cm}\cntDelete\text{-th query} \& \; \& \; \& \dataUnlearnAdd[\cntIteration]\coloneqq\dataUnlearnAdd[\cntIteration]\cup\{(\cntUser,d_{\cntIteration,\cntDelete})\} \& \; 
\\
\; \& \; \& \; \& \; \& \; 
\\
\; \& \; \& \; \& \; \& \; 
\\
\; \& \; \& \; \& \; \& \; 
\\
\hspace*{0.9cm} \pcif \pcnot \VrfyUpdate(\pub,\com[\cntIteration-1],\com[\cntIteration],\proofModel[\cntIteration])  \& \; \& \; \& (\stS[\cntIteration],\model[\cntIteration],\com[\cntIteration],\proofModel[\cntIteration])\gets\ProveUpdate(\stS[\cntIteration-1],\pub,\dataAdd[\cntIteration])~~~ \& \; 
\\
\hspace*{0.9cm}\quad \pcabort \& \; \& \; \& \dataAdd[\cntIteration]\coloneqq \emptyset \& \; 
\\
\; \& \; \& \; \& \; \& \; 
\\
\; \& \; \& \; \& \; \& \; 
\\
\; \& \; \& \; \& \; \& \; 
\\
\hspace*{0.9cm}\pcif \pcnot \VrfyPubUnlearn(\pub,\com[\cntIteration-1],\com[\cntIteration],\proofModel[\cntIteration])\colon \& \; \& \; \& (\stS[\cntIteration],\model[\cntIteration],\com[\cntIteration],\proofModel[\cntIteration]) \gets\ProveUnlearn(\stS[\cntIteration-1],\pub,\dataUnlearnAdd[\cntIteration])\hspace*{0.5cm} \& \; 
\\
\hspace*{0.9cm}\quad \pcabort \& \; \& \; \& \; \& \; 
\\
\; \& \; \& \; \& \pcfor (\cntUser,d_{\cntIteration,\cntDelete})\in\dataUnlearnAdd[\cntIteration]\colon  \& \; 
\\
\hspace*{0.9cm}\pcif \pcnot \VrfyPrivUnlearn(\pub,\cntUser,d_{\cntIteration,\cntDelete},\com[\cntIteration],\proofUnlearn[\cntIteration,\cntDelete])\colon \hspace*{0.5cm} \& \; \& \; \& \quad \proofUnlearn[\cntIteration,\cntDelete]\gets\ProvePrivUnlearn(\stS[\cntIteration],\pub,\cntUser,d_{\cntIteration,\cntDelete})\hspace*{0.5cm} \& \; 
\\
\hspace*{0.9cm}\quad \pcabort \& \; \& \; \& \dataUnlearnAdd[\cntIteration]\coloneqq\emptyset \& \; 
\\
	};
	\draw[dashed] ([xshift=0pt,yshift=0pt] m0-2-1.west) -- ([xshift=-15pt,yshift=0pt] m0-2-5.east) node [draw=none,pos=0.035,below=0cm,black] {\textit{Initialize}};
	\draw[dashed] ([xshift=0pt,yshift=0pt] m0-7-1.west) -- ([xshift=-15pt,yshift=0pt] m0-7-5.east) node [draw=none,pos=0.047,below=0cm,black] {$\cntIteration$-\textit{th iteration}};
	\draw[dashed] ([xshift=20pt,yshift=7pt] m0-15-1.west) -- ([xshift=-15pt,yshift=7pt] m0-15-5.east) node [draw=none,pos=0.063,below=0cm,black] {\textit{Proof of Training}};
    \draw[dashed] ([xshift=20pt,yshift=-2pt] m0-19-1.west) -- ([xshift=-15pt,yshift=-2pt] m0-19-5.east) node [draw=none,pos=0.078,below=0cm,black] {\textit{OR Proof of Unlearning}};
	\draw[>=latex,->] ([xshift=15pt,yshift=0pt] m0-5-3.east) -- ([xshift=-10pt,yshift=0pt] m0-5-2.west) node [draw=none,midway,above=0cm,black] {$\com[0],\proofModel[0]$};
	\draw[>=latex,->] ([xshift=-10pt,yshift=0pt] m0-10-2.west) -- ([xshift=15pt,yshift=0pt] m0-10-3.east) node [draw=none,midway,above=0cm,black] {$\cntUser\in\users,~d_{\cntIteration,\cntAdd}\in\userdata$};
	\draw[>=latex,->] ([xshift=-10pt,yshift=0pt] m0-13-2.west) -- ([xshift=15pt,yshift=0pt] m0-13-3.east) node [draw=none,midway,above=0cm,black] {$\cntUser\in\users,~ d_{\cntIteration,\cntDelete}\in\userdata$};
	\draw[>=latex,->] ([xshift=15pt,yshift=0pt] m0-17-3.east) -- ([xshift=-10pt,yshift=0pt] m0-17-2.west) node [draw=none,midway,above=0cm,black] {$\train\com[\cntIteration],\proofModel[\cntIteration]$};
    \draw[>=latex,->] ([xshift=15pt,yshift=0pt] m0-22-3.east) -- ([xshift=-10pt,yshift=0pt] m0-22-2.west) node [draw=none,midway,above=0cm,black] {$ \unlearn\com[\cntIteration],\proofModel[\cntIteration]$};
     \draw[>=latex,->] ([xshift=15pt,yshift=0pt] m0-25-3.east) -- ([xshift=-10pt,yshift=0pt] m0-25-2.west) node [draw=none,midway,above=0cm,black] {$\proofUnlearn[\cntIteration,\cntDelete]$};
	\coordinate (B) at (current bounding box.south west);
	\draw[line width=0.5pt]
	let
	\p2 = ($(B) - (1mm, -1mm)$)
	in
	(current bounding box.north east) ++(-4mm,-1mm) rectangle (\p2);
	\end{tikzpicture}
}
    	\caption{\emph{Unlearning Framework.} We describe protocols in this framework based on a set of admissible functions $\admissibleFunction$. After initialization, execution proceeds in iterations. In the beginning of each iteration $\cntIteration$, users $\users$ can issue requests for data to be added or deleted. After this phase, the server $\server$ either performs a \emph{proof of training} by adding the requested data records in $\dataAdd[\cntIteration]$ to the model or a \emph{proof of unlearning} by removing the requested data records in $\dataUnlearnAdd[\cntIteration]$. It computes a commitment $\com[\cntIteration]$ on the updated model $\model[\cntIteration]$ and updated training dataset. Furthermore, the server computes a proof $\proofModel[\cntIteration]$ that $\model[\cntIteration]$ was obtained from this dataset. The users verify this proof and the commitment. In each iteration of unlearning the server additionally creates a \emph{proof of non-membership} for every unlearnt data point conforming to a user that it has complied with a data deletion request. This proof can be verified by the user against $\com[\cntIteration]$.}
	\label{fig:protocol-abstract}
    \vspace{-1em}
\end{figure*}

We now present our formal framework for verifiable machine unlearning encompassing the requirements outlined in the previous section. This framework defines verifiable unlearning as an interactive protocol summarized in Figure~\ref{fig:protocol-abstract}.

\mypar{Dataset} Let $\alldata$ be the distribution of data points. Each user $\cntUser\in\users$ possesses a set of data points $\userdata \sim \alldata$.
At the server side, there is an initially empty ``server's dataset'' $\data[0] = \emptyset$ (which is later populated for training an ML model). During the execution of the protocol, users can request to add or delete data points which the server adds respectively deletes from their dataset.
Different versions of the resulting server's dataset (as it develops during the execution of the protocol) are denoted by their corresponding index (\ie $\data[0], \data[1], \dots $).
To attribute data points to users, we assume the server prepends a unique identifier to each data point. Therefore, entries in $\data[\cntIteration]$ are tuples of the form $(\cntUser,d)\in\users\times\userdata$. We refer to such a unique representation as a \emph{data record}.

\mypar{Admissible Functions}\label{sec:admissible-unlearning}
Our approach relies on proving the execution of a {\em pre-specified unlearning algorithm}. 
To capture formally that we cannot consider training and unlearning in isolation, we propose a generic interface that we call \emph{admissible functions}.
One instance is described by a triplet of functions $\admissibleFunction = (\initFunction,\trainingFunction,\allowbreak\unlearningFunction)$, where $\initFunction$ is an initialization function, $\trainingFunction$ is a training function, and $\unlearningFunction$ is an associated unlearning function. The set of all admissible functions is denoted by $\admissibleFunctionsSet$. 
We let $\seed$ denote public hyperparameter which are used for initialization. W.l.o.g., we assume the functions to be deterministic.
Randomness required, \eg for the training, can be obtained deterministically with a pseudo-random number generator and a suitable seed contained in the hyperparameter and stored in the state.
More explicitly:
\begin{itemize}[topsep=2pt, itemsep=\mysep, partopsep=\mysep, parsep=\mysep,leftmargin=15pt]
\item $(\stF,\model)\coloneqq\initFunction(\seed)$: The initialization function $\initFunction$ takes as input hyperparameter $\seed$ and outputs an initial state $\stF$ and model $\model$ (\eg its initial weights).

\item $(\stF,\model)\coloneqq \trainingFunction(\stF,\dataAdd)$: The training function $\trainingFunction$ takes as input the current state $\stF$ and the set $\dataAdd$ of data records to be added. It outputs an updated state and a new model.

\item
$(\stF,\model)\coloneqq \unlearningFunction(\stF,\dataUnlearnAdd)$: The unlearning function $\unlearningFunction$ takes as input the current state $\stF$ and the set $\dataUnlearnAdd$ of data records to be deleted. It outputs an updated state and a new model.
\end{itemize}

These functions allow to establish an abstraction to track the relation between a model and its underlying dataset; we refer to this as the \emph{conceptual} dataset. If $\data$ is the current conceptual dataset, removing data records from $\dataUnlearnAdd$ with $\unlearningFunction$ updates the dataset as $\data \coloneqq \data \setminus \dataUnlearnAdd$.
Before executing the protocol, the server and users must agree on $\admissibleFunction$, similar to the agreement process in the TLS handshake protocol. This agreement must ensure that $\admissibleFunction$ is semantically meaningful and relevant to the context of the application. Just as in the TLS protocol, where the security of the entire protocol depends on selecting a secure cipher suite, the selection of $\admissibleFunction$ is crucial for ensuring the integrity of the process.

\mypar{Protocol Execution} We denote a protocol in our framework by $\protocolF$ for a triplet of functions $\admissibleFunction=(\initFunction,\allowbreak\trainingFunction,\unlearningFunction)\in\admissibleFunctionsSet$. The execution of $\protocolF$ can then be described with two phases (executed in an iterative manner):

\begin{itemize}[topsep=2pt, itemsep=\mysep, partopsep=\mysep, parsep=\mysep,leftmargin=22pt]
\item[{\bf P1.}] {\bf Data Addition/Deletion:} At the beginning of each iteration, user can issue addition/deletion requests to the server. The server batches multiple addition/deletion requests by storing all requests in intermediate datasets $\dataAdd[\cntIteration]$ (addition) and $\dataUnlearnAdd[\cntIteration]$ (deletion).

\item[{\bf P2.}] {\bf Proof of Training (resp.\,Unlearning):} At the end of an iteration $\cntIteration$, the server updates its dataset by adding (resp. deleting) the data records stored in $\dataAdd[\cntIteration]$ (resp.\,$\dataUnlearnAdd[\cntIteration]$). For training (resp.\,unlearning), the server updates the model using function $\trainingFunction$ (resp.\,$\unlearningFunction$) on all records requested to be added (resp. deleted). It then computes a proof of training (resp.\,unlearning) to be verified by the users.
These proofs establishes the state of the evolving dataset across iterations and model updates \emph{and} that unlearnt data records can not be re-added.
Based on the current state, the server then provides each user who requested a point to be deleted with an individual proof that their data record is not part of the dataset (\ie a proof of non-membership).
Finally, at the end of an iteration, the dataset $\dataAdd[\cntIteration]$ (resp.\,$\dataUnlearnAdd[\cntIteration]$) is reset.
\end{itemize}

\subsection{Protocol Syntax}
An unlearning protocol $\protocolF$ w.r.t. admissible functions $\admissibleFunction$ specifies nine algorithms that are executed by the server and users as shown in Figure~\ref{fig:protocol-abstract}.

\mypar{1. Setup and Initialization} A global setup procedure generates public parameters $\pub$, \ie $\pub\gets\Setup(1^\secpar)$, where $\secpar$ is the security parameter. We assume that $\pub$ additionally include functions $\admissibleFunction$ with hyperparameters $\seed$. Depending on the application, this procedure can be executed either by the server or some external entity. Subsequently, $\pub$ is given to all actors. 

During initialization, the ML model and protocol state are initialized using algorithms $\Init$ and $\VrfyInit$. 
Intuitively, the state captures all information required by the protocol (\eg information to keep track of the evolving dataset) as well as information required by $\admissibleFunction$ for training and unlearning. For example, for retraining-based unlearning the state would contain the training dataset $D$, while for amnesiac unlearning, the set of deltas $\parupdate_{\cntepoch, \datasample^*}$ (cf.~\cref{sec:background-ml}) as well as current model $\model$ would be included.
Formally, the following two algorithms are run:

\begin{description}[topsep=2pt, itemsep=\mysep, partopsep=\mysep, parsep=\mysep]
\item[\textbf{Server:}] 

$(\stS[0],\model[0],\com[0],\proofModel[0])\gets\Init(\pub)$

The server initializes $\model[0]$ using initialization function $\initFunction$ and hyperparameter $\seed$ contained in $\pub$. It stores the resulting state $\stF$ in $\stS[0]$. The set of training records is initialized empty, \ie $\data[0] \coloneqq \emptyset$. It then commits to $\model[0]$ and $\data[0]$ with $\com[0] \coloneqq (\com[0]^{\model} \| \com[0]^{\data})$. We assume that the commitment to the initial dataset (and all its updates) is computed deterministically from $\pub$ using a function $\Commit$, \ie $\com[0]^{\data}\coloneqq\Commit(\pub,\data[0])$. 
We do not make additional assumptions about the commitment at this point, but we will later see that it needs to be \emph{binding}. Finally, proof $\proofModel[0]$ attests the initialization of $\model[0]$. 
\item[\textbf{User:}]

$0/1\gets \VrfyInit(\pub,\com[0],\proofModel[0])$

Users verify two things: (a) the commitment $\com[0]$ with $\data[0] = \emptyset$, and (b) the model initialization $\model[0]$ against proof $\proofModel[0]$. If verification is successful, the algorithm outputs $1$. On failure, it outputs $0$.
\end{description}

\mypar{\bf 2A. Proof of Training} In each iteration $i$, where the server performs a proof of training, two algorithms are run:

\begin{description}[topsep=2pt, itemsep=\mysep, partopsep=\mysep, parsep=\mysep]
\item[\textbf{Server:}] 

$(\stS[\cntIteration],\model[\cntIteration],\com[\cntIteration],\proofModel[\cntIteration])$

\hspace*{0.6cm}$\gets\ProveUpdate(\stS[\cntIteration-1],\pub,\dataAdd[\cntIteration])$

The server computes the updated model $\model[\cntIteration]$ by executing training function $\trainingFunction$ on state $\stS[\cntIteration-1]$ and all newly added data records $\dataAdd[\cntIteration]$.
The training set of $\model[\cntIteration]$ is defined as the union of the previous dataset and new data records, \ie $\data[\cntIteration]\coloneqq \data[\cntIteration-1]\cup\dataAdd[\cntIteration]$.
The server commits to both the model and training data with $\com[\cntIteration] \coloneqq (\com[\cntIteration]^{\model} \| \com[\cntIteration]^{\data})$
and computes the proof $\proofModel[\cntIteration]$ that (a) model $\model[\cntIteration]$ was updated by applying $\trainingFunction$,
and (b) training data $\data[\cntIteration]$ does not contain any unlearnt record, \ie $\data[\cntIteration] \cap \dataUnlearn[\cntIteration] = \emptyset$, where $\dataUnlearn[\cntIteration] \coloneqq \bigcup_{k \in [\cntIteration]}{ \dataUnlearnAdd[k]}$ is the set of all unlearnt data records so far. The proof also attests that (c) the set of unlearnt data records has not changed, \ie $\dataUnlearn[\cntIteration-1] = \dataUnlearn[\cntIteration]$. 

\item[\textbf{User:}] 
$0/1\gets \VrfyUpdate(\pub,\com[\cntIteration-1],\com[\cntIteration],\proofModel[\cntIteration])$

Users validate properties (a)-(c) (as described above in $\ProveUpdate$) and the update on commitment $\com[\cntIteration]$ by verifying the proof~$\proofModel[\cntIteration]$ against the previous commitment $\com[\cntIteration-1]$ and the new commitment $\com[\cntIteration]$.
\end{description}

\vspace{1mm}

\noindent{\bf 2B. Proof of Unlearning.} In each iteration $\cntIteration$, where the server performs a proof of unlearning, four algorithms are run:

\begin{description}[topsep=2pt, itemsep=\mysep, partopsep=\mysep, parsep=\mysep]
\item[\textbf{Server:}] 
$(\stS[\cntIteration],\model[\cntIteration],\com[\cntIteration],\proofModel[\cntIteration])$

\hspace*{0.6cm}$\gets\ProveUnlearn(\stS[\cntIteration-1],\pub,\dataUnlearnAdd[\cntIteration])$

The server unlearns all records collected in~$\dataUnlearnAdd[\cntIteration]$ and computes the updated $\model[\cntIteration]$ by executing function~$\unlearningFunction$ on state $\stS[\cntIteration-1]$ and $\dataUnlearnAdd[\cntIteration]$. 
Thus, conceptually, the new training set of~$\model[\cntIteration]$ is defined as $\data[\cntIteration]\coloneqq \data[\cntIteration-1]\setminus\dataUnlearnAdd[\cntIteration]$. Similar to~$\ProveUpdate$, the server commits to both the model and training data with~$\com[\cntIteration]$ and computes the proof~$\proofModel[\cntIteration]$ that (a) model~$\model[\cntIteration]$ was updated by applying~$\unlearningFunction$, and (b) training data~$\data[\cntIteration]$ does not contain any unlearnt records, \ie $\data[\cntIteration] \cap \dataUnlearn[\cntIteration] = \emptyset$, where $\dataUnlearn[\cntIteration] \coloneqq \dataUnlearn[\cntIteration-1] \cup \dataUnlearnAdd[\cntIteration]$. 
The proof also attests that (c) the previous set of unlearnt data records is a subset of the updated set $\dataUnlearn[\cntIteration-1] \subset \dataUnlearn[\cntIteration]$. This ensures that the set of unlearnt data records is append-only and records can never be removed.

\item[\textbf{User:}] 
$0/1 \leftarrow \VrfyPubUnlearn(\pub,\com[\cntIteration-1],\com[\cntIteration],\proofModel[\cntIteration])$

Users validate properties (a)-(c) (as described above in $\ProveUnlearn$) and the update on commitment $\com[\cntIteration]$ by verifying the proof $\proofModel[\cntIteration]$ against the previous commitment $\com[\cntIteration-1]$ and the new commitment $\com[\cntIteration]$.

\item[\textbf{Server:}] 
$\proofUnlearn[\cntIteration,\cntDelete]\gets\ProvePrivUnlearn(\stS[\cntIteration],\pub,\cntUser,d_{\cntIteration,\cntDelete})$

For each record $(\cntUser,d_{\cntIteration,\cntDelete})\in\dataUnlearnAdd[\cntIteration]$, the server computes a proof $\proofUnlearn[\cntIteration,\cntDelete]$ using information from $\stS[\cntIteration]$ that this record is not part of the training set $\model[\cntIteration]$, \ie $(\cntUser,d_{\cntIteration,\cntDelete})\notin \data[\cntIteration]$.

\item[\textbf{User:}] 
$0/1\gets\VrfyPrivUnlearn(\pub,\cntUser,d_{\cntIteration,\cntDelete},\com[\cntIteration],\proofUnlearn[\cntIteration,\cntDelete])$

The user verifies with both $\proofUnlearn[\cntIteration,\cntDelete]$ and $\com[\cntIteration]$ that $(\cntUser,d_{\cntIteration,\cntDelete})$ was not part of the training data $\data[\cntIteration]$ of $\model[\cntIteration]$.
\end{description}

\mypar{Practical considerations}
The framework ensures that once data records are deleted, they cannot be re-added later. Therefore, it suffices if a majority of honest users verify the updates. Even if a user stops participating after confirming their data has been deleted, the verification of updates by the honest majority ensures correct server behavior. In practice, an additional mechanism will be needed for users to report invalid proofs and trigger penalties for the server.

If users trust a third party (such as an auditor), the verification algorithms $\VrfyUpdate$ and $\VrfyPubUnlearn$, can be executed by this entity to minimize redundant computations. The results can then be shared with all users (refer to Section \ref{sec:alt} for further discussion).

Furthermore, it is important to ensure that the server uses the most up-to-date model for inference.
This can be achieved with techniques for verifiable inference~\cite{lee-20-vcnn,liu-21-zkcnn,weng-21-mystique,feng-21-zen,kang-22-scaling}, which is related to, but independent of the problem of unlearning that we consider in this work.

\subsection{Completeness and Security}
\label{sec:completeness}

Within the proposed framework, we can formally describe completeness and security requirements of protocols for verifiable unlearning.

\mypar{Completeness} For completeness, we require that an honest execution of the protocol yields the expected outputs: for an honest server, the users successfully verify the initialization of the model and the proofs for all updates (training and unlearning) performed by the server. A proof of non-membership generated for an unlearnt data record is also successfully verified by the corresponding user. We formally capture this with the following definition.

\begin{definition}[Completeness] Let $\secpar$ be the security parameter. A protocol $\protocolF$ is complete if for all $\pub\gets\Setup(1^\secpar)$, the following properties are satisfied:
\begin{enumerate}[topsep=2pt, itemsep=\mysep, partopsep=\mysep, parsep=\mysep]
\item Let $(\stS[0],\model[0],\com[0],\proofModel[0])\gets\Init(\pub)$. Then $$\Pr[\,\VrfyInit(\pub,\com[0],\proofModel[0])=0\,] \leq \negl(\secpar) \ .$$

\item
Let $\mode[][\cntIteration]\in\{\train[],\unlearn[]\}$ indicate whether proof of training or proof of unlearning has been performed in iteration $\cntIteration$. Let $\advA$ be a PPT adversary that outputs a valid sequence of datasets either to be added  $\{\train\dataAdd[\cntIteration]\}$ or to be deleted $\{\unlearn\dataUnlearnAdd[\cntIteration]\}$ for all $i\in[\noIterations]$. 

For all $i\in[\noIterations]$, if $\mode[][\cntIteration]=\train[]$, let $(\stS[\cntIteration],\model[\cntIteration],\allowbreak\com[\cntIteration],\allowbreak\proofModel[\cntIteration])\gets\ProveUpdate(\stS[\cntIteration-1],\allowbreak\pub,\allowbreak\dataAdd[\cntIteration])$ and if $\mode[][i]=\unlearn[]$, let $(\stS[\cntIteration],\model[\cntIteration],\allowbreak\com[\cntIteration],\proofModel[\cntIteration])\gets\ProveUnlearn(\stS[\cntIteration-1],\allowbreak\pub,\allowbreak\dataUnlearnAdd[\cntIteration])$.

Then for all $\mode[][i]=\train$
\vspace*{1mm}

{\centering
\hspace*{-0cm}\scalebox{0.92}{
$\Pr[\VrfyUpdate(\pub,\com[\cntIteration-1],\com[\cntIteration],\proofModel[\cntIteration])=0]\leq\negl(\secpar) \ ,$
}\vspace*{2mm}}
and for all $\mode[][i]=\unlearn$
\vspace*{1mm}

{\centering
\hspace*{-0cm}\scalebox{0.91}{
$\Pr[\VrfyPubUnlearn(\pub,\com[\cntIteration-1],\com[\cntIteration],\proofModel[\cntIteration])=0]\leq\negl(\secpar) \ ,$
}\vspace*{2mm}}
where validity is defined via the following conditions: $\forall i,j$ s.\,t. $i\neq j\colon \dataAdd[i]\cap\dataAdd[j]=\emptyset$ and $\forall i,j$ s.\,t. $j< i\colon\dataAdd[i]\cap\dataUnlearnAdd[j]=\emptyset$.
    
\item For all $\cntIteration\in[\noIterations]$ s.\,t. $\mode[][\cntIteration]=\unlearn[]$: for all $(u,d)\in\dataUnlearnAdd[\cntIteration]$, let $\proofUnlearn\gets\ProvePrivUnlearn\allowbreak(\stS[\cntIteration], \pub,\cntUser,d)$, then\vspace*{1mm}

{\centering
\hspace*{-0.2cm}\scalebox{0.9}{
$\Pr[\VrfyPrivUnlearn(\pub,u,d,\com[\cntIteration],\proofUnlearn)=0]\leq\negl(\secpar) \;\!.$
}\vspace*{2mm}}
\end{enumerate}
\label{def:completeness}
\end{definition}

In this definition, we only require \emph{computational} completeness to allow for a wide range of instantiations. For example, an instantiation that works on hash values of data records cannot achieve perfect completeness because of hash collisions. By allowing for computational completeness, however, we only require that it should be hard for a PPT adversary to find such collisions (\ie with a negligible probability).

\mypar{Security} The security of an unlearning protocol can be modelled in terms of a security game. In this game, the adversary, described by a probabilistic algorithm $\advA$, takes the role of the server. Intuitively, the definition need to capture that a malicious server cannot add (and train on) a data record that a user requested to delete in a previous iteration. 
Formally this is modeled as the winning condition in the security game in Figure \ref{fig:sec-unlearn}.
To this end, we define an \emph{extractability-based} security model; this allows to cover realistic attackers that only need to output valid transcripts of the interactions, while the extractability property ensures that the server must know some underlying dataset for these transcripts. Note that similar security definition are commonly used in the context of hash functions or SNARKs~\cite{bitansky-12-extractable,fiore-16-hash,groth-16-size}.

The adversary in our game has to provide the protocol outputs (\ie the commitments and proofs), whereas the extractor outputs the corresponding inputs that the adversary used (\ie the underlying datasets). We give a formal description of game $\SecurityUnlearn$ in \cref{fig:sec-unlearn}, which is divided into two stages:

\begin{itemize}[topsep=2pt, itemsep=\mysep, partopsep=\mysep, parsep=\mysep]
    \item[{\bf S1.}] \textbf{Simulation.} The game draws the public parameters $\pub$ using $\Setup$ and runs the adversary $\advA$ on input $\pub$. The extractor $\mathcal{E}$ is run on the same input and random coins.
    Additionally, we provide auxiliary input $\aux$  which captures any extra information the adversary may have whenever the protocol is used in combination with other cryptographic schemes and allows for a wide range of possible instantiations. As commonly done, we restrict this input to only \emph{benign} inputs~\cite{bitansky-16-extractable,bitansky-17-hunting}, e.g., we do not allow the auxiliary input to encode an arbitrary (possibly obfuscated) circuit.
    At some point, $\advA$ will terminate and output a sequence of tuples $(k,(u,d),\proofUnlearn,\{\mode[\textnormal{:}~][\cntIteration]\com[\cntIteration],\proofModel[\cntIteration]\}_{\cntIteration\in[0,\noIterations]})$ for some $\noIterations\in\mathbb{N}$, where $(u,d)$ is a data record that was proved to be deleted in the $k$-th iteration, and $\mode[][\cntIteration]\in\{\train[],\unlearn[]\}$. At the same time, the extractor outputs a sequence of datasets $(\data[0],\ldots,\data[\noIterations])$.
    
    \item[{\bf S2.}] \textbf{Finalize.} After the adversary has terminated, the game uses the extractor's output to compute the set of data records unlearnt in the $k$-th iteration based on the datasets $\data[k]$ and $\data[k-1]$. Recall that the commitment in the framework consists of two parts $\com[\cntIteration]^{\model}$ and $\com[\cntIteration]^{\data}$, where we need the second part for verification. The game checks for the following conditions: (a) $\com[\cntIteration]^{\data}$ was obtained from $\data[\cntIteration]$, (b) the initial proof $\proofModel[0]$ verifies for the initial commitment $\com[0]$, (c) each proof of training $\proofModel[\cntIteration]$ verifies for commitments $\com[\cntIteration-1]$ and $\com[\cntIteration]$, (d) each proof of unlearning $\proofModel[\cntIteration]$ verifies for commitments $\com[\cntIteration-1]$ and $\com[\cntIteration]$, (e) the proof of non-membership $\proofUnlearn$ verifies for $(\cntUser,d)$ and $\com[k]$, (f) $k<\noIterations$ and $(\cntUser,d)$ was unlearnt in iteration $k$ and re-added in iteration $\noIterations$. If all these properties are satisfied, then the game outputs $1$ and $\advA$ wins.
\end{itemize}

We summarize this with the following definition.
\nicoresetlinenr
\begin{figure}[t]
	\centering
	\hspace*{-0.25cm}
	\scalebox{0.85}{
		\fbox{\small
			\begin{minipage}[t]{9.75cm}
			    \underline{$\SecurityUnlearn_{\advA,\mathcal{E},\protocolF,\alldata}(1^\secpar)$}
			    \begin{nicodemus}
			        \item $\pub\gets\Setup(1^\secpar)$
			        \item $(k,(u,d),\proofUnlearn,\{\mode[\textnormal{:}~][\cntIteration]\com[\cntIteration],\proofModel[\cntIteration]\}_{\cntIteration\in[0:\noIterations]} ;~ $
                    \item[] \hspace*{3.8cm}$\{\data[\cntIteration]\}_{\cntIteration\in[0:\noIterations]}) \gets(\advA\|\mathcal{E})(\pub,\aux)$
			        \medskip
			         \item \codecomment{Pre-processing}
                    \item $\dataUnlearnAdd[k]\coloneqq\data[k-1]\setminus\data[k]$
			        \item Parse $\com[\cntIteration]$ as $(\com[\cntIteration]^{\model}\|\com[\cntIteration]^{\data})~\forall\cntIteration\in[0:\noIterations]$
                    \medskip
			        \item \codecomment{Evaluate winning condition}
			        \item \pcif $\Commit(\pub,\data[\cntIteration])=\com[\cntIteration]^{\data} ~\forall\cntIteration\in[0:\noIterations]$ \hfill \codecomment{Datasets\!}
			         \item ~~$\pcand \VrfyInit(\pub,\com[0],\proofModel[0])$ \hfill \codecomment{Initialization\!}
			         \item ~~$\pcand \VrfyUpdate(\pub,\com[\cntIteration-1],\com[\cntIteration],\proofModel[\cntIteration])~~$
                    \item[] \hspace*{4.5cm} $\forall\cntIteration:\mode[][\cntIteration]=\train[]$ \hfill \codecomment{Training\!}
                    \item ~~$\pcand \VrfyPubUnlearn(\pub,\com[\cntIteration-1],\com[\cntIteration],\proofModel[\cntIteration])~$
                    \item[] \hspace*{4.5cm} $\forall\cntIteration:\mode[][\cntIteration]=\unlearn[]$ \hfill \codecomment{Unlearning\!}
                    \item ~~$\pcand \VrfyPrivUnlearn(\pub,u,d,\com[k],\proofUnlearn)$ \hfill \codecomment{Non-Membership\!}
                    \item ~~$\pcand (u,d)\in \dataUnlearnAdd[k]$ \hfill \codecomment{Point unlearnt}
                    \item ~~$\pcand (u,d)\in\data[\noIterations] ~\pcand k< \noIterations \colon$ \hfill \codecomment{Point re-added later\!}
                    \item \quad \pcreturn $1$
			        \item \pcreturn $0$
			    \end{nicodemus}
			\end{minipage}
	}}
	\caption{
		\emph{Security Game.} We define the security of an protocol $\protocolF$ in terms of game $\SecurityUnlearn$. The notation $(\advA\|\extractor)$ denotes that both algorithms are run on the same input and random coins and assigning their results to variables before resp.~after the semicolon. Input $\aux$ refers to auxiliary input.
		\label{fig:sec-unlearn}}
  \vspace{-1em}
\end{figure}
\begin{definition}[Unlearning]\label{def:security}
Let $\secpar$ be the security parameter and consider game~$\SecurityUnlearn$ in \cref{fig:sec-unlearn}. Protocol~$\protocolF$ for data distribution~$\alldata$ is unlearning-secure if for all PPT adversaries~$\advA$ there exists an extractor~$\extractor$ such that for all benign auxiliary inputs~$\aux\colon$
$$\Pr[\SecurityUnlearn_{\advA,\extractor,\protocolF,\alldata}(1^\secpar)\Rightarrow1]\leq\negl(\secpar) \, .$$
\end{definition}
\section{Instantiation}
\label{sec:instantiation}

Our framework defines a generic interface for constructing protocols for verifiable unlearning.
To instantiate a protocol within this framework, we need to address two main challenges. First, we need to be able to verify the correct execution of the training and unlearning algorithms to validate any changes to the dataset. Second, we require a mechanism to keep track of the dataset across iterations, enabling queries to verify (non\mbox{-})membership of specific data records.

In this section, we introduce the core components to account for these challenges and present a practical instantiation of an unlearning protocol based on SNARKs and hash functions. 
Our instantiation is generic and we prove its completeness and security universally for any triplet $(\initFunction,\trainingFunction,\unlearningFunction)$ of admissible functions. 
An overview of the full protocol is depicted in Appendix~\ref{app:protocol}.

\mypar{Data Representation} To represent the dataset, we split all data records as those belonging to either training data $\data$ or unlearnt data $\dataUnlearn$.
For our instantiation, the server stores two ordered sets $\hashsData$ and $\hashsUnlearned$ of hashed training data records and unlearnt data records. From both sets, we additionally compute a hash value in the form of a hash chain (cf. $\HashData$ in Appendix \ref{app:protocol}). 
This representation allows for efficient caching of intermediate hashes and, for~$\hashsUnlearned$, enables us to easily prove that entries are append-only (i.e., prevent records from being removed from the chain) as well as fast membership verification for unlearnt data records. To account for the partition of training and unlearnt data as well as the user admissible function, we instantiate the commitment~$\com$ as a tuple of four elements: hash of (a) the state~$\hStF$ (defined by~$\admissibleFunction$), (b) the model~$\hModel$, (c) the training data~$\hData$, and (d) the unlearnt data~$\hUnlearn$. Looking ahead, a collision-resistant hash function is sufficient for the binding property; it ensures that the adversary cannot come up with a second input that has the same hash value.

\mypar{Proof System} To verify the correct execution of~$\initFunction$, $\trainingFunction$, and~$\unlearningFunction$, we use proof systems. More specifically, SNARKs, which allow (broadly speaking) to prove statements of the form that an output $y$ is the result of applying a function $f$ on an input $x$, i.e., $y \coloneqq f(x)$.
Therefore, we define the verification of the initialization, training updates and unlearning updates in terms of polynomial decidable binary relations $\relationI$, $\relationT$ and $\relationU$ over circuits $\circuitInit$, $\circuitTraining$ and $\circuitUnlearning$ (resp.) as introduced in Section \ref{subsec:crypto_primitives}. These circuits describe the required computations---based on $\initFunction$, $\trainingFunction$, and~$\unlearningFunction$. We exemplary outline $\circuitUnlearning$ in Figure~\ref{fig:circuits-u}; for circuit $\circuitInit$, $\circuitTraining$ refer to Appendix~\ref{app:circuits}.
Note that for verification, only public parameters are required.
Furthermore, by using SNARKs, we can keep the instantiation generic and universally prove its completeness and security for any triplet $(\initFunction,\trainingFunction,\unlearningFunction)$.
\nicoresetlinenr
\begin{figure}[t]
	\centering
	\hspace*{-0.15cm}\scalebox{0.77}{
		\fbox{\small
			\begin{minipage}[t]{10.05cm}
               \underline{$\circuitUnlearning(\textbf{public } \hStF[,\cntIteration],\hStF[,\cntIteration-1],\hModel[\cntIteration],\hData[\cntIteration],\hData[\cntIteration-1],\hUnlearn[\cntIteration],\hUnlearn[\cntIteration-1],$}\\
                \hspace*{0.45cm}\underline{$\textbf{ private } \stF[,\cntIteration-1], \hashsData[\cntIteration-1],\dataUnlearnAdd[\cntIteration])$~}
				\begin{nicodemus}
                    \item \codecomment{Check input set of hashed training data records}
                    \item \pcif $\hData[\cntIteration-1]\neq\HashData(\hashsData[\cntIteration-1])\colon$
                    \item \quad \hspace*{0.1cm}$\pcreturn \false$
                    \item \vspace{0.3em}\codecomment{Update and check set of hashed unlearnt data records and training data records}
                    \item $\hashsUnlearnedAdd[\cntIteration]\coloneqq \{\HashDataSample(\cntUser,d)\}_{(\cntUser,d)\in\dataUnlearnAdd[\cntIteration]}$
                    \item $\hashsData[\cntIteration]\coloneqq\hashsData[\cntIteration-1] \setminus \hashsUnlearnedAdd[\cntIteration]$
                    \item $\pcif \hUnlearn[\cntIteration]\neq\AppendHashDataUnlearn(\hUnlearn[\cntIteration-1],\hashsUnlearnedAdd[\cntIteration])$ \pcor
                    $\hData[\cntIteration]\neq\HashData(\hashsData[\cntIteration])\colon$
                    \item \quad \hspace*{0.1cm}$\pcreturn \false$
                    \item \vspace{0.3em} \codecomment{Check input state, perform unlearning and check outputs}
                    \item $\hStF[,\cntIteration-1]\neq\HashState(\stF[,\cntIteration-1])\colon$
                    \item \quad \pcreturn $\false$
                    \item $(\stF[,\cntIteration],\model[\cntIteration])\coloneqq\unlearningFunction(\stF[,\cntIteration-1],\dataUnlearnAdd[\cntIteration])$
                    \item \pcif $\hStF[,\cntIteration]\neq\HashState(\stF[,\cntIteration])$ \pcor
                    $\hModel[\cntIteration]\neq\HashModel(\model[\cntIteration])\colon$
                    \item \quad \hspace*{0.1cm}$\pcreturn \false$
                    \item \vspace{0.3em} $\pcreturn \true$
				\end{nicodemus}
\end{minipage}}}
	\caption{\emph{Circuits $\circuitUnlearning$.} Based on the circuit, we prove correct execution of admissible functions for the proof of unlearning. 
	\label{fig:circuits-u}}
 \vspace{-2.5em}
\end{figure}

\mypar{1. Initialization} During the protocol's initialization, function $\initFunction$ is run to obtain the initial state $\stF[,0]$ and initial model $\model[0]$. Also, the sets of hashed training data and unlearnt data records are initialized, \ie $\hashsData[0]=\emptyset$ and $\hashsUnlearned[0]=\emptyset$. The commitment consists of hashes to these four values, \ie $\com[0]=(\hStF[,0],\hModel[0],\hData[0],\hUnlearn[0])$. Correct initialization is proved using the SNARK for relation $\relationI$ captured by circuit $\circuitInit$. The proof of training $\proofModel[0]$ consists of the statement $\statement_0$ and resulting SNARK proof $\proofI[0]$, which can be verified by the user using $\com[0]$.

\mypar{2A. Proof of Training} The server starts by executing $\ProveUpdate$. In the $\cntIteration$-th iteration, it first performs the model update by running function $\trainingFunction$ on the previous state $\stF[,\cntIteration-1]$ and new data records $\dataAdd[\cntIteration]$, the result being an updated state $\stF[,\cntIteration]$ and a new model $\model[\cntIteration]$. Then the server updates the set of hashed training data records $\hashsData[\cntIteration]$ with $\dataAdd[\cntIteration]$ and computes the new commitment $\com[\cntIteration]=(\hStF[,\cntIteration],\hModel[\cntIteration],\hData[\cntIteration],\hUnlearn[\cntIteration-1])$, where the commitment to the unlearnt data records is the same as in the previous iteration since no data was deleted.

The proof $\proofModel[\cntIteration]$ is computed using the SNARK for relation $\relationT$ captured by circuit $\circuitTraining$. The corresponding statement $\statement_\cntIteration$ and proof $\proofT[\cntIteration]$ attest that (a) the model and state were updated correctly with $\dataAdd[\cntIteration]$, (b) the set of hashed unlearnt data was not changed, and (c) no data record that was previously unlearnt is added. The server sends $(\proofModel[\cntIteration],\com[\cntIteration])$ to the users. 
Subsequently, the users execute $\VrfyUpdate$ and verify $\proofModel[\cntIteration]$ using $\com[\cntIteration]$ and the previous commitment $\com[\cntIteration-1]$.

\mypar{2B. Proof of Unlearning} The proof of unlearning consists of two parts: the model update for deleting data records and the proof of non-membership. The server first runs~$\ProveUnlearn$. In the~$\cntIteration$-th iteration, it performs the model update by running function~$\unlearningFunction$ on the previous state~$\stF[,\cntIteration-1]$ and the set~$\dataUnlearnAdd[\cntIteration]$ of data records to be deleted. The result is the updated state~$\stF[,\cntIteration]$ and model~$\model[\cntIteration]$. The set~$\hashsUnlearned[\cntIteration]$ is computed by appending hashed records of~$\dataUnlearnAdd[\cntIteration]$ to~$\hashsUnlearned[\cntIteration-1]$. At the same time, $\hashsData[\cntIteration]$ is computed from~$\hashsData[\cntIteration-1]$ by removing those entries. The commitment~$\com[\cntIteration]$ consists of the hash values~$(\hStF[,\cntIteration],\hModel[\cntIteration],\hData[\cntIteration],\hUnlearn[\cntIteration])$. The whole procedure is proved using circuit~$\circuitUnlearning$ for relation~$\relationU$, producing a SNARK proof~$\proofU[\cntIteration]$ for the corresponding statement~$\statement_\cntIteration$, which can be verified by the user using~$\com[\cntIteration]$ and~$\com[\cntIteration-1]$.

The second part of the proof of unlearning is to provide a proof on non-membership to all users that requested a data record~$(\cntUser,\datasample)\in\dataUnlearnAdd[\cntIteration]$ to be deleted. We prove this by proving its membership in~$\hashsUnlearned[\cntIteration]$. If~$\hashsUnlearned[\cntIteration] \cap \hashsData[\cntIteration] = \emptyset$, it follows that $(\cntUser,\datasample) \notin \data[\cntIteration]$ (which we show to hold when proving completeness). %
Specifically, we use the hash chain for~$\hashsUnlearned[\cntIteration]$: for a data record, we compute a membership proof as a path in this chain; this path can be verified by recomputing the chain and comparing the final result with the hash in the commitment (\ie hash value~$\hashDataUnlearn[\cntIteration]$).

Thus, the server performs~$\ProvePrivUnlearn$ by computing the chain path to a data record~$(\cntUser,\datasample) \in \dataUnlearnAdd[\cntIteration]$. It outputs this as proof $\proofUnlearn$ which is sent to the user. The user uses the hash~$\hUnlearn[\cntIteration]$ from the commitment to verify membership. If the path leads to that hash, the user accepts, and aborts otherwise.

\subsection{Completeness and Security}
\label{sec:proof}

We now want to prove the completeness and security of the instantiated protocol. We start to show that our instantiation is complete according to \cref{def:completeness}. We give a sketch in the following; refer to Appendix \ref{sec:completeness-proof} for the full proof.

\begin{theorem}\label{thm:completeness}
Let $\SNARK$ be a complete SNARK and $\Hash$ a collision-resistant hash function. Then the instantiated protocol satisfies completeness.
\end{theorem}

\begin{proof}[Proof (Sketch)]
Completeness of the initialization (first property) is easy to observe since the two hashed datasets are initialized as empty and the execution of function $\initFunction$ is proven with the SNARK for relation $\relationI$. By completeness of the SNARK, the users can successfully verify the proof, additionally using the commitments to state, model and datasets. The second property follows from the completeness of the SNARKs for relations $\relationT$ and $\relationU$ and collision-resistance of the hash function. However, note that if a hash collision occurs, it is not possible to provide the proof of training. Thus, only computational completeness can be achieved. Given that the proofs of training and unlearning are successful, completeness of the proof of non-membership (third property) follows from the construction and correctness of the hash chain.
\end{proof}

We continue to prove that the instantiation is a secure unlearning protocol according to \cref{def:security}. We again give a sketch below; refer to Appendix~\ref{sec:security-proof} for the full proof.

\begin{theorem}\label{thm:security}
Let $\Hash$ be a collision-resistant hash function and $\SNARK$ be a secure SNARK. Then the instantiated protocol satisfies unlearning security.
\end{theorem}

\begin{proof}[Proof (Sketch)]
Let $\advA$ be an adversary in the unlearning security game (cf.\,\cref{fig:sec-unlearn}). By knowledge soundness of the SNARK, there exists an extractor which outputs the witness and thus the datasets $\data[\cntIteration]$ corresponding to the outputs of the adversary. We then use the soundness of the SNARK. That is, $\advA$ must have computed the proof using a witness, \ie the state $\stF[,\cntIteration]$ and the dataset $\dataAdd[\cntIteration]$ (in the proof of training) or dataset $\dataUnlearnAdd[\cntIteration]$ (in the proof of unlearning), which also determine the model $\model[\cntIteration]$ and must correspond to the hash values in the commitment. By collision-resistance of the hash function, the adversary cannot find another state, model or dataset for the same commitment. Thus, applying function $\trainingFunction$ (or $\unlearningFunction$) to the previous state and datasets results in same state and model as used by $\advA$.

Since all proofs as well as the proof of non-membership of data record $(u,d)$ must verify successfully, the hash of $(u,d)$ must be contained in the set $\hashsUnlearned[k]$ which was used to create the proof. Here, $k$ is the iteration where $(u,d)$ was unlearnt; the observation holds by assuming soundness of the SNARK and collision-resistance of the hash function. We can further infer that $(u,d)$ must also be part of future sets $\hashsUnlearned[\cntIteration]$, $k<i\leq \noIterations$ and by collision-resistance $(u,d)$ must also be part of the underlying datasets $\dataUnlearn[\cntIteration]$. Finally, we use the fact that the proof attests that the intersection of $\hashsUnlearned[\noIterations]$ and $\hashsData[\noIterations]$ is empty. This yields a contradiction and shows that $(\cntUser,d)$ cannot be present in the last dataset $\data[\noIterations]$.
\end{proof}

\section{Implementation}
\label{sec:implementation}

Next, we implement and compare the main building blocks of the instantiated protocol. For our implementation, we consider different triplets of functions $\admissibleFunction = (\initFunction,\trainingFunction,\unlearningFunction)$ based on  techniques from the machine unlearning literature; namely, retraining-based unlearning, amnesiac unlearning and optimization-based unlearning (cf.~\cref{sec:background-ml}). Additionally, we study the applicability to different ML models and datasets.

To practical implement the protocol, we first need to instantiate SNARK $\SNARK$ and hash function $\Hash$, and then define circuits $\circuitInit$, $\circuitTraining$ and $\circuitUnlearning$ for all three sets of functions $\admissibleFunction$. 
Our code is available at \href{https://github.com/cleverhans-lab/verifiable-unlearning}{http://github.com/verifiable-unlearning/artifacts}. Experiments are performed on a server running Ubuntu 22.04 with 256 GB RAM and two Intel Xeon Gold 5320 CPUs.

\mypar{Proof System} Our instantiation is generic and can be implemented with any secure SNARK that satisfies completeness, soundness, and knowledge soundness (cf. Section \ref{subsec:crypto_primitives}). In this work, we use Spartan~\cite{setty-20-spartan} as it is efficient and, more importantly, transparent, \ie it does not require a trusted setup. Spartan comes in two variants, as a succinct non-interactive zero-knowledge (NIZK) proof system and as a SNARK. Similar to the work of Angel~\emph{et al.}~\cite{angel-22-efficient}, we use the NIZK variant, where verification time is linear in the size of the R1CS instance (see below). By using the SNARK variant, some verification cost can be offset to the server and a one-time pre-processing step for the user.
Depending on the application, it may be beneficial to use a different proof system. One alternative is Groth16~\cite{groth-16-size}, which, for example, requires a trusted setup, but has the advantage of constant verification time and proof size.

\mypar{Circuits}
Spartan is implemented on the \verb|ristretto255| elliptic curve, a prime-order group abstraction atop \verb|curve25519|.
Following prior work on verifiable computation~\cite{angel-22-efficient, ozdemir-20-scaling, wu-18-dizk}, we convert the computation of our circuits into \ac{R1CS} instances; \ie the statements in $\relationI$, $\relationT$ and $\relationU$ (cf. \cref{fig:circuits-u} and Appendix~\ref{app:circuits}) are represented as a constraint system over a {\em finite field}.
More specifically, an R1CS instance is described by a tuple $(\mathbb{F},A,B,C,io,n)$, where $\mathbb{F}$ is the finite field, $A,B,C\in\mathbb{F}^{n\times n}$ are matrices of size $n\geq |io|+1$ and $io$ is the public input and output of the instance. R1CS is a generalization of arithmetic circuit satisfiability. We say an R1CS instance is satisfiable if there exists a witness $\omega\in\mathbb{F}^{n-|io|-1}$ such that $(A\cdot z)\circ(B\cdot z)=(C\cdot z)$ for $z=(io, 1, \omega)$, where $\cdot$ is the matrix-vector product and $\circ$ the Hadamard product. Since $A,B,C$ are generally sparse matrices, a parameter $n$ is sometimes specified, denoting the maximum number of non-zero entries in each matrix.

We implement the algorithms for training and unlearning as arithmetic circuits using the ZoKrates programming language~\cite{eberhardt-18-zokrates} and use CirC~\cite{ozdemir-22-circ} for compilation into \ac{R1CS} instances.
To represent data and other parameters in a finite field, we convert them into fixed precision real numbers.

\mypar{Hash Function} We only require collision-resistance for the hash function.
Although our instantiation is generic and can work with any hash function, it is beneficial to use an algebraic hash function where most operations can be directly done in the finite field of the SNARK.
Bit-wise hash functions such as the SHA family of hash functions are much slower in that regard.
For our construction, we use Poseidon~\cite{grassi-21-poseidon} as it is particularly designed for zero-knowledge proof systems. Other good options include Pedersen Hash~\cite[p.76]{misc-zcash} or MIMC~\cite{albrecht-16-mimc}. 
To be used with Spartan, we implement a version of Poseidon for the \verb|ristretto255| curve.

\begin{table}[t]
\centering
\caption{\emph{Run-Time of Protocol Functions.} We compare the running time between the protocols subtasks. We consider retraining-based unlearning, amnesiac unlearning, and optimization-based unlearning. We report the relative difference with retraining in gray.}
\label{table:protocol-stages}
\resizebox{\columnwidth}{!}{
\begin{tabular}{@{}l r >{\color{gray}}l r >{\color{gray}}l r >{\color{gray}}l @{}}
\toprule
& \multicolumn{2}{@{}c}{{{\bf Retraining}} } 
& \multicolumn{2}{@{}c}{{{\bf Amnesiac}} } 
& \multicolumn{2}{@{}c}{{{\bf Optimization}}} \\
\midrule
\midrule
\multicolumn{4}{@{}l}{\emph{Proof of Training}} \\
\; R1CS & 8,056,887 & $\times 1.00$ & 8,130,535 & $\times 1.01$ & 7,980,878 & $\times 0.99$ \\
\rule{0pt}{2ex}
\; $\SNARKProve $ w/ $\relationT$ &  4m 32s & $\times 1.00$ &  4m 32s & $\times 1.00$ &  4m 31s & $\times 0.99$ \\
\rule{0pt}{2ex}
\; $\SNARKVrfy$  w/ $\relationT$ &  1m 36s & $\times 1.00$ &  1m 37s & $\times 1.01$ &  1m 35s & $\times 0.99$ \\[0.5ex]
\midrule 
\multicolumn{4}{@{}l}{\emph{Proof of Unlearning}} \\
\; R1CS & 8,102,288 & $\times 1.00$ & 616,005 & $\times 0.08$ & 919,456 & $\times 0.11$ \\
\rule{0pt}{2ex}
\; $\SNARKProve $ w/ $\relationU$ &  4m 58s & $\times 1.00$ &  2m 18s & $\times 0.46$ &  0m 53s & $\times 0.18$ \\
\rule{0pt}{2ex}
\; $\SNARKVrfy$  w/ $\relationU$ &  1m 48s & $\times 1.00$ &  0m 49s & $\times 0.45$ &  0m 20s & $\times 0.19$ \\[0.5ex]
\midrule 
\multicolumn{4}{@{}l}{\emph{Proof of Non-Membership}} \\
\rule{0pt}{2ex}
\;  $\ComputeTreePath$  & \phantom{xxm~x}$<1$s & $\times 1.00$  
                        & \phantom{xxm~x}$<1$s & $\times 1.00$
                        & \phantom{xxm~x}$<1$s & $\times 1.00$ \\
\rule{0pt}{2ex}
\; $\VrfyTreePath$  & \phantom{xxm~x}$<1$s & $\times 1.00$  
                    & \phantom{xxm~x}$<1$s & $\times 1.00$
                    & \phantom{xxm~x}$<1$s & $\times 1.00$ \\[0.5ex]              
\bottomrule
\end{tabular}}
\begin{flushleft}
\scriptsize{{R1CS}: \#constraints}\\
\vspace{-2.5em}
\end{flushleft}
\end{table}
\subsection{Protocol Functions}
\label{sec:instantiation-admissible-algos} 

We start by comparing the subtasks of proof of training, proof of unlearning, and proof of non-membership between the different sets of unlearning algorithms. 
Therefore, we implement the high-level functions of the instantiated protocol for retraining-based unlearning, amnesiac unlearning~\cite{graves-21-amnesiac}, and optimization-based unlearning~\cite{jang-22-knowledge, warnecke-21-unlearning}.
The main focus is on the comparison as well as to show feasibility and versatility of our approach. We discuss possible improvements, \eg regarding scalability, in Section \ref{sec:scalability}.

First, we want to compare and understand the overheads of each subtask between techniques. To this end, we consider a linear regression model and train this model for 3 epochs with SGD as a general purpose approach. We use a synthetic dataset $\data$ and set the batch size to 1. 
We compute a proof of training with the addition of 100 data points with 10 features each. We set $|\data[0]| = 0$, $|\dataAdd[1]| = 100$, and $|\dataUnlearn[0]| = 0$ accordingly. Subsequently, we compute the proof of unlearning and simulate the deletion of 10 data points and set $|\data[1]| = 100$, $|\dataUnlearn[1]| = 0$, and $|\dataUnlearnAdd[2]| = 10$. For optimization-based unlearning, we unlearn for 3 epochs. The results are presented in Table \ref{table:protocol-stages}.
The complexity of a statement for Spartan can be measured as the numbers of R1CS constraints. 
Across all techniques, compilation time of R1CS instances ranges between 17s (optimization-based) and 48m 45s (retraining-based).

\mypar{Proof of Training}
We observe that the complexity of the training is comparable between unlearning approaches. The underlying R1CS instances have between $7,980,878$---$8,130,535$ constraints with proving time between 4m~31s---4m~32s. Recall that in amnesiac unlearning, we also need to collect model updates that are later used for unlearning, which introduces negligible overhead compared to the training costs. 

\mypar{Proof of Unlearning}
Runtime of generating and verifying the proof of unlearning shows more variance. Amnesiac unlearning is over $2\times$ faster and optimization-based unlearning over $5\times$ faster than retraining-based unlearning.
This is despite the R1CS instance of optimization-based unlearning being almost 50\% larger compared to the amnesiac instance (919,456 vs. 616,005 constraints) but it is still more efficient to compute as it is 63\% more sparse (\ie 7,660,455 vs. 12,248,390 entries are non-zero).
The main difference is that amnesiac unlearning requires to maintain and verify a state from training (\ie the model updates) while optimization-based unlearning does not require a state. 

\mypar{Proof System}
In general, we observe that verification is $2\times$--$3\times$ faster than proof generation. This is dependent on the choice of the proof system. For example, by using the SNARK variant of Spartan, we can offload some of the verification costs to the server and an additional pre-processing for the user. In this case, proving time increases to 33m~39s---34m~12s for the proof of training across all techniques and verification time reduces to $<1$s, but the user needs to run a one-time pre-processing step which takes between 8m~6s---8m~12s.

\mypar{Proof of Non-Membership}
Finally, proof of non-membership is highly efficient. The implementation is independent of the unlearning scheme, and both proof generation and verification take less than one second.
\begin{table}[t]
\centering
\caption{\emph{Proving Time vs. Model Capacity.} We compare the proving time of proof of training for different classes of models with increasing capacity.}
\label{table:classifier}
\resizebox{\columnwidth}{!}{
\begin{tabular}{@{}lrrrrrr@{}}
\toprule
{{\bf Classifier}} 
&& \multicolumn{1}{@{}c}{ {{\bf R1CS}} } 
&& \multicolumn{1}{@{}c}{ $\boldsymbol{\SNARKProve}$ }
&& \multicolumn{1}{@{}c}{ $\boldsymbol{\SNARKVrfy}$ } \\
\midrule
\midrule
Linear Regression && 8,056,887 &&  4m 33s &&  1m 36s \\
\rule{0pt}{2ex}
Logistic Regression && 9,048,909 &&  5m  8s &&  1m 45s \\
\rule{0pt}{2ex}
Neural Network ($N=2$) && 21,867,010 &&  9m 50s &&  3m 34s \\
\rule{0pt}{2ex}
Neural Network ($N=4$) && 42,030,731 && 24m 16s &&  6m 42s \\
\bottomrule
\end{tabular}}
\begin{flushleft}
\scriptsize{{R1CS}: \#constraints}\\
\end{flushleft}
\vspace{-2.5em}
\end{table}

\subsection{Model and Dataset Complexity}
The dominant component of the protocols' run-time is the complexity of the circuit used to generate proofs of training and unlearning. This complexity depends mainly on (a) the unlearning technique, (b) the complexity of the model, and (c) the size of the dataset.
In the following, we first consider model complexity and study different classes of models. Next, we look on the complexity of the dataset. In both cases, we focus on retraining-based unlearning as the baseline from Table \ref{table:protocol-stages} and, more specifically, on the training circuit $\circuitTraining$.

\mypar{Models}
To understand the effects of the choice of ML model, we follow related work~\cite{zhao-21-veriml}, and consider linear regression, logistic regression and neural networks for classification. For the neural networks, we focus on models with one hidden layer and varying numbers of (hidden) neurons $N \in \{2, 4\}$. For activation, we use the sigmoid function and approximate it with a third-order polynomial as done in~\cite{kim-18-logistic, kim-18-secure}.
Again, we train each model with SGD for 3 epochs on a synthetic dataset consisting of 100 training points with 10 features each. 

The results are summarized in \cref{table:classifier}. We observe that the R1CS instance increases together with the complexity of the model. For example, the number of constraints increases by $1.12\times$ to $9,048,909$ constraints when going from linear to logistic regression. This is intuitive: in logistic regression, we additionally need to evaluate the sigmoid activation which induces this overhead. In a similar vein, moving from logistic regression to neural networks increases the circuit further to $21,867,010$ ($N=2$) and $42,030,731$ ($N=4$) constraints.

\mypar{Benchmark Datasets}
To understand the impact of the dataset, we choose several datasets from the PMLB benchmark suite~\cite{romano-21-pmlb} (as considered in related work~\cite{angel-22-efficient} on verifiable computation of numerical optimization problems) and train a linear regression model for all datasets. 
To make results comparable, we train all models for 3 epochs with a learning rate of 0.1. As commonly done, we split the data into 80:20 train test split. Models achieve a test accuracy between 73\,\% and 92\,\%.

Results are presented in Appendix \ref{app:scalability-datasets}.
For all models, we observe a linear dependence between run-time and dataset size. Generating a proof for the smallest dataset with 600 total features (\ie total points $\times$ features) requires 2m~22s and for the largest dataset with 3,324 total features requires 13m~36s.

\section{Discussion}
\label{sec:discussion}

Next, we discuss alternative ways to instantiate protocols within our framework and improvements to our construction.

\subsection{Alternative Instantiations}
\label{sec:alt}
\mypar{External Trust} 
Our approach eliminates the need for a trusted third party by relying solely on cryptographic protocols to ensure security. To enhance efficiency and reduce the burden on users, a trusted auditor can be introduced to verify on their behalf (as we discuss towards the end of \cref{sec:framework-overview}). This could be done using either a dedicated trusted third party, such as another cloud provider with no incentive to collude with the server, or by employing distributed auditors where trust derives from independent verifications. For instance, Meta recently announced that they will implement key transparency in WhatsApp to complement the manual scanning of QR codes, outsourcing verification to a third party~\cite{meta-key-trans}.

\mypar{Trusted Hardware} If TEEs (\eg Intel SGX~\cite{mckeen-13-innovative}) are available, training and unlearning procedures can be performed within them. More specifically, we can replace the SNARK in our instantiation with a TEE  such that the circuit is computed inside the TEE. In this case, the proof consists of a digest signed by the TEE provider, which can be verified by the user. However, when using a TEE, one needs to consider common concerns such as trusting a hardware vendor, availability of said vendor for signing the digests, limited memory~\cite{grover-18-privado}, their applicability to ML-related tasks that involve GPU computation~\cite{volos-18-graviton}, and side-channels~\cite{ohrimenko-16-oblivious,wang-17-leaky}. Some of these issues have been addressed by Weng \etal \cite{weng-22-proof} (cf.~\cref{sec:related-work}).

\mypar{Minimizing Redundancy} In our instantiation, users who request unlearning are required to verify future updates to ensure that their data has not been reintroduced. If we combine VC with an additional proof of secure data erasure, we can give similar guarantees while not requiring the user to verify all updates. However, secure erasure is a non-trivial problem in itself and was considered in \eg~\cite{perito-10-secure}. Formalizing deletion compliance from a server's perspective \cite{garg-20-data} can also be seen as complementary problem.

\subsection{Scalability}
\label{sec:scalability}
Our experiments show that the run-time of the instantiated protocol is dominated by generating and verifying the proof of training and unlearning. We base our construction on \ac{VC} and, as a result, inherit its scalability limitations. This can also be observed for other VC-based approaches in the ML setting~\cite{shafran-21-crypto, helbitz-21-reducing, kim-18-logistic, kim-18-secure}. Any future advances in \ac{VC} will lead to run-time improvements for our approach. Nevertheless, we want discuss how we can improve performance with the primitives available today. 

\mypar{SNARK-friendly Techniques}
Certain computations are more amendable to efficient SNARK verification than others, for example, the development of SNARK-friendly hash functions~\cite{grassi-21-poseidon, albrecht-16-mimc, misc-zcash}. 
Similarly, there exist ML paradigms that are also more amenable to verification. For instance, inference using quantized models~\cite{kang-22-scaling, feng-21-zen} or lookup tables for expensive computations~\cite{kang-22-scaling} reduce costs. Furthermore, when there exists a unique ML model (\ie a global optimum for the underlying optimization problem), proving and verification complexity can be improved even further \cite{angel-22-efficient}. 
In our experiments, we observe that online computation of model updates in optimization-based unlearning is faster than verifying model updates in amnesiac unlearning as the verification of input values involves expensive calculation of hash values.
We envision future work to focus on developing SNARK-friendly unlearning techniques combining above observations.

\mypar{Offloading Computation}
Orthogonal to the employed unlearning technique and ML model, one can offload expensive proof generation steps to the user (\eg the evaluation of a non-linear activation function).
We can split the proving and verification processes such that the server creates a proof for certain types of computations and shares partial results with the (honest) user who performs (and thus verifies) expensive computations themselves.

\mypar{Application-specific Relaxations}
Finally, depending on the application, it might be possible to avoid the expensive generation of the proof of training.
Consider, for instance, an application where data is collected once and {\em only} be removed at a later point in time (\eg biomedical user studies or other human-involved data collection processes).
In this case, proof of training only needs to be performed once and---if users further trust the initial training phase---it might be sufficient to only prove unlearning. 

\subsection{Privacy}
Formalizing privacy for unlearning protocols is an interesting direction for future work and requires to establish an additional security definition. Although it is out-of-scope for our work, we want to highlight that our instantiation does not require the users to know the datasets or model. In fact, they only see hash commitments and the SNARK proofs. If the hash function satisfies pseudo-randomness or is modeled as a random oracle, and  assuming that the input space is large enough, then hash values do not leak information about the underlying data points. Additionally, if the SNARK satisfies the zero-knowledge property~\cite{goldwasser-89-knowledge} (which most SNARKs including Spartan do), then the proof also does not leak information about the witness. However, we require users to know whether training or unlearning happened because they need to know which verification procedure to run. Privacy in the context of model inference has been studied more extensively, \eg Gao \etal \cite{gao-22-deletion} define security notions for deletion hiding and reconstruction. An overview for different formalizations of inference privacy is also given by Salem \etal \cite{salem-22-privacy}.

\section{Related Work}\label{sec:related-work}
Our approach to verifiable unlearning intersects with various areas of security and ML research. Below, we explore related concepts and methods.

\mypar{Verifying Unlearning} 
Prior work~\cite{gao-22-verifi,sommer-20-towards} aims at verifying unlearning by embedding backdoors~\cite{gu-17-badnets} in models (using data whose unlearning is to be verified) and verifying backdoor removal on unlearning. However, such approaches are probabilistic with no theoretical guarantees of when they work, unlike our cryptography-informed approach which produces verifiable proofs. 
The work of Guo \etal~\cite{guo-19-certified} provides end-users with a certificate that the new model is influenced by the specific data in a quantifiably low manner. While this certificate conceptually bounds the influence of a data point from an algorithmic perspective, it provides no guarantee that the entity executing the algorithm (\ie server) did so correctly. In our work, we aim to capture exactly this and provide cryptographic guarantees. %
Weng \etal \cite{weng-22-proof} propose an unlearning framework based on TEEs. 
Their protocol uses SISA unlearning~\cite{bourtoule-19-machine} and can be captured by our framework as well. 
In contrast to our instantiation that is based only on cryptographic primitives, their approach relies on trusted hardware (\ie the correctness and integrity of the SGX enclave) as well as cryptographic assumptions (\ie EUF-CMA security of the signature scheme used by the enclave and collision-resistance of the hash function). 

\mypar{Verifiable Computation}
We use verifiable computation for proof of training and proof of unlearning. 
There has been a series of works demonstrating a remarkable progress in making these schemes (and those related to verification of data used for computation) practical~(\eg\cite{setty-20-spartan,gennaro-10-noniteractive,parno-13-pinocchio,fiore-16-hash}). 
To verify the computation of training an ML model, Zhao \etal~\cite{zhao-21-veriml} also propose verification using a SNARK. However, their primary objective is to design a scheme to ensure that the payments made to servers are correct. In our work, we design a scheme to verify the correctness of data deletion when training ML models.
Otti~\cite{angel-22-efficient} is a compiler that is aimed at designing efficient arithmetic circuits for problems that involve optimization (such as those commonly found in ML). DIZK~\cite{wu-18-dizk} is a distributed system capable of distributing the compute required for proof creation.
Garg \etal~\cite{garg-23-experimenting} describes an approach to verify ML training based on MPC-in-the-head, while the work of Abbaszadeh \etal~\cite{abbaszadeh-24-kaizen} developes a sumcheck-based proof system for the gradient-descent algorithm and framework for recursive composition of proofs. Both approaches may be extended to a full-fledged (retraining-based) unlearning protocol using our framework.

\mypar{Proving Model Inference}
There exist various approaches to proving inference using SNARKs \cite{lee-20-vcnn,liu-21-zkcnn,weng-21-mystique,feng-21-zen,kang-22-scaling} which complements our protocol in that regard. Another approach would be to use trusted execution environments to do so as suggested in \cite{weng-22-proof}.
\section{Conclusion}
The problem of unlearning has gained significant interest in terms of definitions and algorithms for updating model parameters. However, regardless of the definition or the algorithm the server uses to update the model, the user has no way to verify that the server indeed executed the unlearning procedure. In this paper, we define unlearning as a security problem and propose a framework to capture the guarantees verifiable unlearning needs to provide. We propose the first verifiable unlearning procedure based on cryptographic primitives instantiated using SNARKs and hash chains. Our implementation shows the feasibility of our approach on several benchmark datasets and machine learning models. Future work includes determining which unlearning techniques are most suitable for efficient verifiable computation, while at the same time devising methods specifically for verifying machine learning code.
\section*{Acknowledgements}

This work was supported by the Deutsche Forschungsgemeinschaft (DFG, German Research Foundation) under the project ALISON (492020528) and Germany’s Excellence Strategy - EXC 2092 CASA - 390781972. Thorsten Eisenhofer conducted this research while interning at the Vector Institute.  Olga Ohrimenko has been supported in part by the joint CATCH MURI-AUSMURI. Nicolas Papernot would like to acknowledge his sponsors, who support his research with financial and in-kind contributions, including Apple, CIFAR through the Canada CIFAR AI Chair program, DARPA through the GARD program, Intel, NSERC through the Discovery grant, Meta, and Ontario through the Early Researcher Award program. Resources used in preparing this research were provided, in part, by the Province of Ontario, the Government of Canada through CIFAR, and companies sponsoring the Vector Institute. We would like to thank members of the CleverHans Lab for their feedback. We would also like to thank Sebastian Angel, Jess Woods, and Eleftherios Ioannidis for input related to the SNARK compilers.
\bibliographystyle{plain}
\bibliography{strings, main}

\clearpage
\appendices
\onecolumn
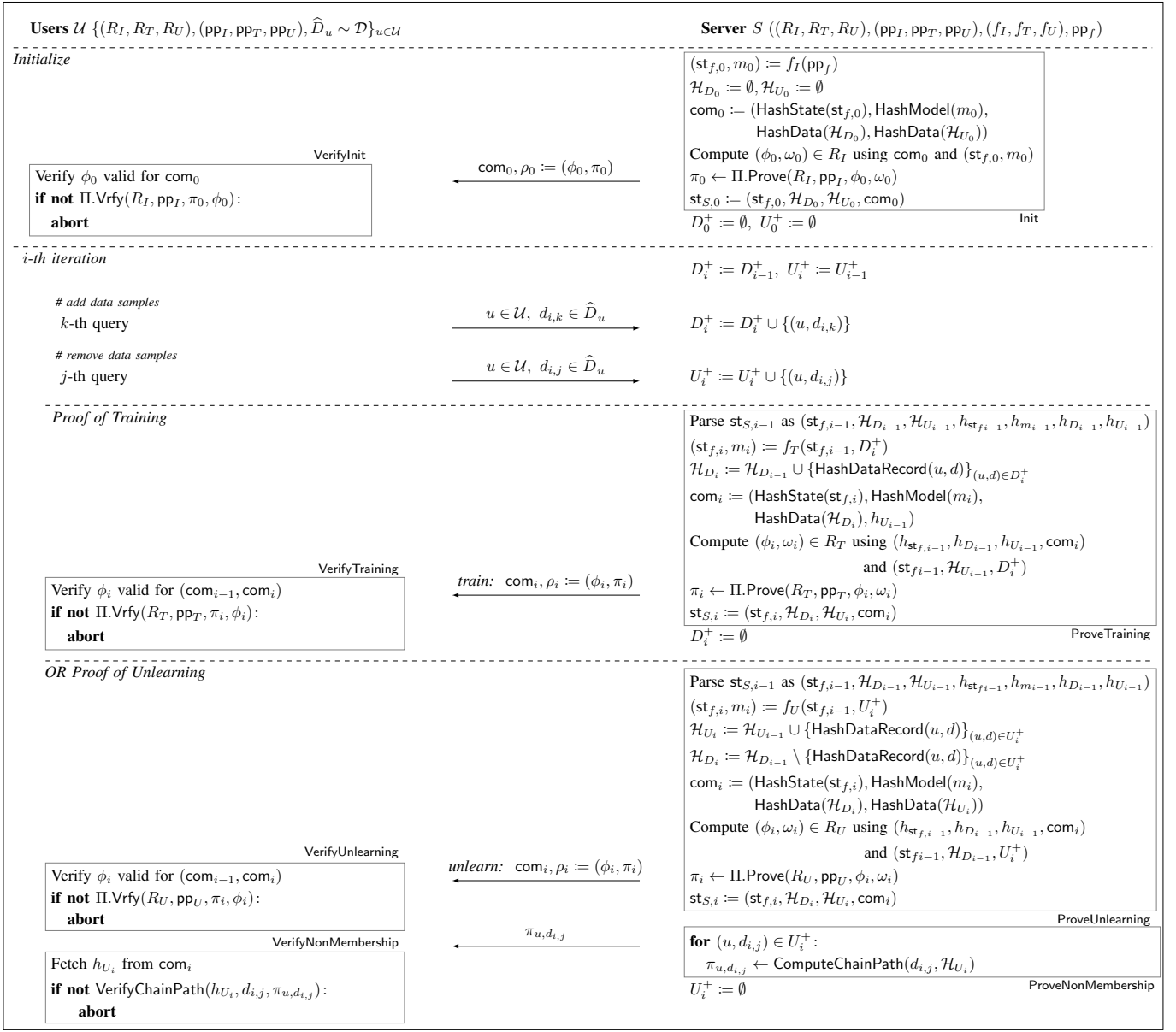
\section{Additional Algorithms and Full Protocol}
\label{app:protocol}

We instantiate the protocol $\protocolF$ for the triple of admissible functions $\admissibleFunction=(\initFunction,\trainingFunction,\unlearningFunction)$ with two primitives: a SNARK $\SNARK$, and a hash function $\Hash$. 
Additional algorithms are described below.

\begin{center}
\hspace*{-0.05cm}\resizebox{1.02\textwidth}{!}{
	\begin{tikzpicture}
	\matrix (m0) [matrix of math nodes,row sep=-0.3em,column sep=0em,minimum width=1em,minimum height=1em,ampersand replacement=\&,
	column 1/.style={anchor=base west},
	column 4/.style={anchor=base west},
	column 7/.style={anchor=base west}] {
~~\textbf{Users}~\users~ \{(\relationI,\relationT,\relationU),(\ppI,\ppT,\ppU),\userdata\sim\alldata\}_{\cntUser\in\users} \& \hspace*{2.8cm} \& \hspace*{2.8cm} \& ~~\textbf{Server}~\server~((\relationI,\relationT,\relationU),(\ppI,\ppT,\ppU),(\initFunction,\trainingFunction,\unlearningFunction),\seed) \& \;
\\
\; \& \; \& \; \& \; \& \;
\\
\; \& \; \& \; \& (\stF[,0],\model[0])\coloneqq\initFunction(\seed)
\& \; 
\\
\; \& \; \& \; \& \hashsData[0]\coloneqq \emptyset, \hashsUnlearned[0]\coloneqq \emptyset \& \; 
\\
\; \& \; \& \; \& \com[0]\coloneqq (\HashState(\stF[,0]),\HashModel(\model[0]),\& \;
\\
\; \& \; \& \; \& \phantom{\com[0]\coloneqq(}\HashData(\hashsData[0]),\HashDataUnlearn(\hashsUnlearned[0]))  \& \;
\\
\; \& \; \& \; \& \text{Compute } (\statement_0,\witness_0)\in\relationI \text{ using } \com[0] \text{ and } (\stF[,0],\model[0]) \& \;
\\
\quad \text{Verify } \statement_0 \text{ valid for } \com[0] \& \; \& \; \& \proofI[0]\gets\SNARKProve(\relationI,\ppI,\statement_0,\witness_0) \& \; 
\\
\quad \pcif \pcnot \SNARKVrfy(\relationI,\ppI,\proofI[0],\statement_0)\colon \& \; \& \; \& \stS[0]\coloneqq(\stF[,0],\hashsData[0],\hashsUnlearned[0],\com[0]) \& \;
\\
\quad \quad \pcabort \& \; \& \; \& \dataAdd[0]\coloneqq\emptyset,~ \dataUnlearnAdd[0]\coloneqq\emptyset \& \;
\\
\; \& \; \& \; \& \; \& \; 
\\
\; \& \; \& \; \& \; \& \; 
\\
\; \& \; \& \; \& \dataAdd[\cntIteration]\coloneqq\dataAdd[\cntIteration-1],~ \dataUnlearnAdd[\cntIteration]\coloneqq\dataUnlearnAdd[\cntIteration-1] \& \;
\\
\; \& \; \& \; \& \; \& \;
\\
\hspace*{0.8cm}\codecomment{add data samples} \& \; \& \; \& \; \& \;
\\
\hspace*{0.9cm}\cntAdd\text{-th query} \& \; \& \; \& \dataAdd[\cntIteration]\coloneqq\dataAdd[\cntIteration]\cup\{(\cntUser,d_{\cntIteration,\cntAdd})\} \& \;
\\
\; \& \; \& \; \& \; \& \; 
\\
\hspace*{0.8cm}\codecomment{remove data samples} \& \; \& \; \& \; \& \; 
\\
\hspace*{0.9cm}\cntDelete\text{-th query} \& \; \& \; \& \dataUnlearnAdd[\cntIteration]\coloneqq\dataUnlearnAdd[\cntIteration]\cup\{(u,d_{\cntIteration,\cntDelete})\} \& \;
\\
\; \& \; \& \; \& \; \& \; 
\\
\; \& \; \& \; \& \; \& \; 
\\
\; \& \; \& \; \& \text{Parse } \stS[\cntIteration-1] \text{ as } (\stF[,\cntIteration-1],\hashsData[\cntIteration-1],\hashsUnlearned[\cntIteration-1],\hStF[\cntIteration-1],\hModel[\cntIteration-1],\hData[\cntIteration-1],\hUnlearn[\cntIteration-1]) \; \& \; 
\\
\; \& \; \& \; \& (\stF[,\cntIteration],\model[\cntIteration])\coloneqq\trainingFunction(\stF[,\cntIteration-1],\dataAdd[\cntIteration]) \& \;
\\
\; \& \; \& \; \& \hashsData[\cntIteration]\coloneqq \hashsData[\cntIteration-1] \cup \{\HashDataSample(\cntUser,d)\}_{(\cntUser,d)\in\dataAdd[\cntIteration]} \& \; 
\\
\; \& \; \& \; \& \com[\cntIteration]\coloneqq (\HashState(\stF[,\cntIteration]),\HashModel(\model[\cntIteration]), \& \;
\\
\; \& \; \& \; \& \phantom{\com[\cntIteration]\coloneqq(}\HashData(\hashsData[\cntIteration]),\hUnlearn[\cntIteration-1]) \& \; 
\\
\; \& \; \& \; \& \text{Compute $(\statement_{\cntIteration},\witness_{\cntIteration}) \in \relationT$ using } (\hStF[,\cntIteration-1],\hData[\cntIteration-1],\hUnlearn[\cntIteration-1],\com[\cntIteration]) \& \;
\\
\; \& \; \& \; \& \phantom{\text{Compute  }(\statement_{\cntIteration},\witness_{\cntIteration}) \in \relationT~} \text{~ and }(\stF[\cntIteration-1],\hashsUnlearned[\cntIteration-1],\dataAdd[\cntIteration]) \& \;
\\
\qquad \text{Verify } \statement_\cntIteration \text{ valid for } (\com[\cntIteration-1],\com[\cntIteration])  \& \; \& \; \& \proofT[\cntIteration]\gets\SNARKProve(\relationT,\ppT,\statement_{\cntIteration},\witness_{\cntIteration})  \& \;
\\
\qquad \pcif \pcnot \SNARKVrfy(\relationT,\ppT,\proofT[\cntIteration],\statement_\cntIteration)\colon \& \; \& \; \& \stS[\cntIteration]\coloneqq(\stF[,\cntIteration],\hashsData[\cntIteration],\hashsUnlearned[\cntIteration],\com[\cntIteration]) \& \; 
\\
\qquad \quad \pcabort \& \; \& \; \& \dataAdd[\cntIteration]\coloneqq\emptyset \& \; 
\\
\; \& \; \& \; \& \; \& \; 
\\
\; \& \; \& \; \& \; \& \; 
\\
\; \& \; \& \; \& \text{Parse } \stS[\cntIteration-1] \text{ as } (\stF[,\cntIteration-1],\hashsData[\cntIteration-1],\hashsUnlearned[\cntIteration-1],\hStF[\cntIteration-1],\hModel[\cntIteration-1],\hData[\cntIteration-1],\hUnlearn[\cntIteration-1]) \& \; 
\\
\; \& \; \& \; \&  (\stF[,\cntIteration],\model[\cntIteration])\coloneqq\unlearningFunction(\stF[,\cntIteration-1],\dataUnlearnAdd[\cntIteration]) \& \;
\\
\; \& \; \& \; \& \hashsUnlearned[\cntIteration]\coloneqq \hashsUnlearned[\cntIteration-1] \cup \{\HashDataSample(\cntUser,d)\}_{(\cntUser,d)\in\dataUnlearnAdd[\cntIteration]} \& \;
\\
\; \& \; \& \; \& \hashsData[\cntIteration]\coloneqq \hashsData[\cntIteration-1] \setminus \{\HashDataSample(\cntUser,d)\}_{(\cntUser,d)\in\dataUnlearnAdd[\cntIteration]} \& \;
\\
\; \& \; \& \; \& \com[\cntIteration]\coloneqq (\HashState(\stF[,\cntIteration]),\HashModel(\model[\cntIteration]), \& \;
\\
\; \& \; \& \; \& \phantom{\com[\cntIteration]\coloneqq(}\HashData(\hashsData[\cntIteration]),\HashDataUnlearn(\hashsUnlearned[\cntIteration])) \& \; 
\\
\;  \& \; \& \; \& \text{Compute $(\statement_{\cntIteration},\witness_{\cntIteration}) \in \relationU$ using } (\hStF[,\cntIteration-1],\hData[\cntIteration-1],\hUnlearn[\cntIteration-1],\com[\cntIteration]) \& \;
\\
\; \& \; \& \; \& \phantom{\text{Compute  }(\statement_{\cntIteration},\witness_{\cntIteration}) \in \relationU~} \text{~ and }(\stF[\cntIteration-1],\hashsData[\cntIteration-1],\dataUnlearnAdd[\cntIteration]) \& \;
\\
\qquad \text{Verify } \statement_\cntIteration \text{ valid for } (\com[\cntIteration-1],\com[\cntIteration])  \& \; \& \; \& \proofU[\cntIteration]\gets\SNARKProve(\relationU,\ppU,\statement_{\cntIteration},\witness_{\cntIteration}) \& \;
\\
\qquad \pcif \pcnot \SNARKVrfy(\relationU,\ppU,\proofT[\cntIteration],\statement_\cntIteration)\colon  \& \; \& \; \& \stS[\cntIteration]\coloneqq(\stF[,\cntIteration],\hashsData[\cntIteration],\hashsUnlearned[\cntIteration],\com[\cntIteration]) \& \;
\\
\qquad \quad \pcabort \& \; \& \; \& \; \& \; 
\\
\; \& \; \& \; \&  \pcfor (\cntUser,d_{\cntIteration,\cntDelete})\in \dataUnlearnAdd[\cntIteration]\colon  \& \;
\\
\qquad \text{Fetch } \hUnlearn[\cntIteration] \text{ from } \com[\cntIteration]  \& \; \& \; \& \quad \proofUnlearn[\cntIteration,\cntDelete]\gets\ComputeTreePath(\dUnlearn[\cntIteration,\cntDelete],\hashsUnlearned[\cntIteration]) \& \;
\\
\qquad \pcif \pcnot \VrfyTreePath(\hUnlearn[\cntIteration],\dUnlearn[\cntIteration,\cntDelete],\proofUnlearn[\cntIteration,\cntDelete])\colon \& \; \& \; \& \dataUnlearnAdd[\cntIteration]\coloneqq\emptyset \& \; 
\\
\qquad \quad ~~\pcabort \& \; \& \; \& \; \& \; 
\\
	};
	\draw[dashed] ([xshift=0pt,yshift=0pt] m0-2-1.west) -- ([xshift=-15pt,yshift=0pt] m0-2-5.east) node [draw=none,pos=0.025,below=0cm,black] {\textit{Initialize}};
	\draw[dashed] ([xshift=0pt,yshift=3pt] m0-12-1.west) -- ([xshift=-15pt,yshift=3pt] m0-12-5.east) node [draw=none,pos=0.045,below=0cm,black] {$\cntIteration$-\textit{th iteration}};
	\draw[dashed] ([xshift=20pt,yshift=0pt] m0-21-1.west) -- ([xshift=-15pt,yshift=0pt] m0-21-5.east) node [draw=none,pos=0.058,below=0cm,black] {\textit{Proof of Training}};
    \draw[dashed] ([xshift=20pt,yshift=3pt] m0-33-1.west) -- ([xshift=-15pt,yshift=3pt] m0-33-5.east) node [draw=none,pos=0.072,below=0cm,black] {\textit{OR Proof of Unlearning}};
    \draw[>=latex,->] ([xshift=10pt,yshift=0pt] m0-8-3.east) -- ([xshift=-10pt,yshift=0pt] m0-8-2.west) node [draw=none,midway,above=0cm,black] {$\com[0],\proofModel[0]\coloneqq(\statement_0,\proofI[0])$};
	\draw[>=latex,->] ([xshift=-10pt,yshift=0pt] m0-16-2.west) -- ([xshift=10pt,yshift=0pt] m0-16-3.east) node [draw=none,midway,above=0cm,black] {$\cntUser\in\users,~\dUnlearn[\cntIteration,\cntAdd]\in\userdata$};
	\draw[>=latex,->] ([xshift=-10pt,yshift=0pt] m0-19-2.west) -- ([xshift=10pt,yshift=0pt] m0-19-3.east) node [draw=none,midway,above=0cm,black] {$\cntUser\in\users,~\dUnlearn[\cntIteration,\cntDelete]\in\userdata$};
	\draw[>=latex,->] ([xshift=10pt,yshift=0pt] m0-29-3.east) -- ([xshift=-10pt,yshift=0pt] m0-29-2.west) node [draw=none,midway,above=0cm,black] {$\train\com[\cntIteration],\proofModel[\cntIteration]\coloneqq(\statement_{\cntIteration},\proofT[\cntIteration])$};
    \draw[>=latex,->] ([xshift=10pt,yshift=0pt] m0-42-3.east) -- ([xshift=-10pt,yshift=0pt] m0-42-2.west) node [draw=none,midway,above=0cm,black] {$\unlearn\com[\cntIteration],\proofModel[\cntIteration]\coloneqq(\statement_{\cntIteration},\proofU[\cntIteration])$};
    \draw[>=latex,->] ([xshift=10pt,yshift=0pt] m0-45-3.east) -- ([xshift=-10pt,yshift=0pt] m0-45-2.west) node [draw=none,midway,above=0cm,black] {$\proofUnlearn[\cntIteration,\cntDelete]$};
	\draw[draw=gray] ([xshift=0pt,yshift=-1pt] m0-3-4.north west) rectangle ([xshift=83pt,yshift=2pt] m0-9-4.south east) node [draw=none,black,below=-0.05cm,anchor=north east] {\footnotesize{$\Init$}};
	\draw[draw=gray] ([xshift=10pt,yshift=0pt] m0-8-1.north west) rectangle ([xshift=-60pt,yshift=0pt] m0-10-2.south west) node [draw=none,black,above=2cm,anchor=north east] {\footnotesize{$\VrfyInit$}};
    \draw[draw=gray] ([xshift=0pt,yshift=-1pt] m0-22-4.north west) rectangle ([xshift=-2pt,yshift=0pt] m0-30-5.south west)node [draw=none,black,below=-0.05cm,anchor=north east] {\footnotesize{$\ProveUpdate$}};
	\draw[draw=gray] ([xshift=20pt,yshift=0pt] m0-29-1.north west) rectangle ([xshift=-50pt,yshift=0pt] m0-31-2.south east) node [draw=none,black,above=2cm,anchor=north east] {\footnotesize{$\VrfyUpdate$}};
	\draw[draw=gray] ([xshift=0pt,yshift=0pt] m0-34-4.north west) rectangle ([xshift=-2pt,yshift=0pt] m0-43-5.south west) node [draw=none,black,below=-0.05cm,anchor=north east] {\footnotesize{$\ProveUnlearn$}};
    \draw[draw=gray] ([xshift=20pt,yshift=0pt] m0-42-1.north west) rectangle ([xshift=-50pt,yshift=1pt] m0-44-2.south east) node [draw=none,black,above=1.95cm,anchor=north east] {\footnotesize{$\VrfyPubUnlearn$}};
	\draw[draw=gray] ([xshift=0pt,yshift=0pt] m0-45-4.north west) rectangle ([xshift=-10pt,yshift=0pt] m0-46-5.south east) node [draw=none,black,below=-0.05cm,anchor=north east] {\footnotesize{$\ProvePrivUnlearn$}};
	\draw[draw=gray] ([xshift=20pt,yshift=0pt] m0-46-1.north west) rectangle ([xshift=-50pt,yshift=1pt] m0-48-2.south east) node [draw=none,black,above=2cm,anchor=north east] {\footnotesize{$\VrfyPrivUnlearn$}};
	\coordinate (B) at (current bounding box.south west);
	\draw[line width=0.5pt]
	let
	\p2 = ($(B) - (1mm, 0mm)$)
	in
	(current bounding box.north east) ++(-4mm,1mm) rectangle (\p2);
	\end{tikzpicture}}
 \end{center}

\begin{center}
    \nicoresetlinenr
	\hspace*{-0.1cm}\scalebox{.833}{
		\fbox{\small
			\begin{minipage}[t]{4cm}
                \underline{$\HashData(\hashs)$}
				\begin{nicodemus}
                    \item $\treeroot \coloneqq \Hash(\dataempty)$
                    \item $\pcfor \hashDataSample\in\hashs\colon$
                    \item \quad $\treeroot \coloneqq \Hash(\treeroot, \hashDataSample)$
				    \item $\pcreturn \treeroot$
				\end{nicodemus}
				\medskip
                \underline{$\AppendHashData(h,\hashs)$}
				\begin{nicodemus}
                    \item $\treeroot \coloneqq h$
                    \item $\pcfor \hashDataSample\in\hashs\colon$
                    \item \quad $\treeroot \coloneqq \Hash(\treeroot, \hashDataSample)$
				    \item $\pcreturn \treeroot$
				\end{nicodemus}
                \medskip
			\end{minipage}
   			\begin{minipage}[t]{5cm}
                      \underline{$\HashDataSample(\cntUser,d=(\x, \y))$}
				\begin{nicodemus}
                    \item $\hashDataSample \coloneqq \Hash(\cntUser)$
                    \item $\pcfor \x_j \in\x\colon$
                    \item \quad $\hashDataSample \coloneqq \Hash(\hashDataSample, \Hash(\x_j))$
                    \item $\hashDataSample \coloneqq \Hash(\hashDataSample, \Hash(\y))$
				    \item \pcreturn $\hashDataSample$
				\end{nicodemus}
                \medskip
				\underline{$\HashModel(\model = [w_0, \dots, w_n])$}
				\begin{nicodemus}
                    \item $\hModel \coloneqq \Hash(m[0])$
                    \item $\pcfor w_i \in m[\text{1:\phantom{1}}] \colon$
                    \item \quad $\hModel \coloneqq \Hash(\hModel, \Hash(w_i))$
				    \item \pcreturn $\hModel$
				\end{nicodemus}
			\end{minipage}
   			\begin{minipage}[t]{6.5cm}
      				\underline{$\HashState(\stF)$}
				\begin{nicodemus}
                    \item $\hStF \coloneqq \Hash(\stF[][0])$
                    \item $\pcfor s_i \in\hStF[][\text{1:\phantom{1}}]\colon$
                    \item \quad $\hStF \coloneqq \Hash(\hStF, s_i)$
				    \item $\pcreturn \hStF$
				\end{nicodemus}
                    \medskip
                    \underline{$\VrfyTreePath(\cntUser,\dUnlearn,\hashDataUnlearn,\proofUnlearn)$}
				\begin{nicodemus}
                    \item \codecomment{recompute hash $\treeroot$ from path $\proofUnlearn$}
				    \item $\treeroot \coloneqq \Hash(\proofUnlearn[][0], \HashDataSample(\cntUser,\dUnlearn))$
                    \item $\pcforin{\text{node}}{\proofUnlearn[][1\colon\,]}$
                    \item \quad $\treeroot \coloneqq \Hash(\treeroot, \text{node})$
                    \item \codecomment{verify final hash}
                    \item $\pcreturn \bool{\treeroot = \hashDataUnlearn}$
				\end{nicodemus}	
                \medskip
			\end{minipage}
   			\begin{minipage}[t]{5.6cm}
      \underline{$\ComputeTreePath(\cntUser,\dUnlearn,\hashsUnlearned)$}
				\begin{nicodemus}
                    \item $h_{\dUnlearn} \coloneqq \HashDataSample(\cntUser,\dUnlearn)$
				    \item $\idx_{\dUnlearn} \coloneqq \hashsUnlearned.\idxf(h_{\dUnlearn})$
				    \item $\pcif \idx_{\dUnlearn}=\bot\colon$
				    \item \quad $\pcreturn \bot$
                    \item \codecomment{get intermediate hash from chain below $\dUnlearn$}
                    \item $\treeroot \coloneqq \HashDataUnlearn(\hashsUnlearned[\,\colon\idx_{\dUnlearn}])$
                    \item $\proofUnlearn \coloneqq [ \treeroot ]$
                    \item \codecomment{add path from  $\dUnlearn$}
                    \item $\pcfor \hashDataSample \in \hashsUnlearned[\idx_{\dUnlearn}+1\colon\,]\colon$
                    \item \quad $\proofUnlearn.\append(\hashDataSample)$
                    \item $\pcreturn \proofUnlearn$
				\end{nicodemus}
			\end{minipage}
   }}\!\!
			\vspace*{0.5cm}
\end{center}

\section{Circuits}
\label{app:circuits}
Based on circuits $\circuitInit$ and $\circuitTraining$, we prove correct execution of admissible functions for initialization and proof of training.
\begin{center}
\nicoresetlinenr
	\hspace*{-0.15cm}\scalebox{0.83}{
		\fbox{\small
			\begin{minipage}[t]{8cm}
   				\underline{$\circuitInit(\textbf{public } \hStF[,0],\hModel[0],\hData[0],\hUnlearn[0], \textbf{ private } \stF[,0], \model[0]$)~}
				\begin{nicodemus}
                    \item \codecomment{Check input for initialization}
                    \item \pcif $\hStF[,0]\neq\HashState(\stF[,0])$ \pcor
                    \item[] \phantom{\pcif}$\hModel[0]\neq\HashModel(\model[0])$ \pcor
                    \item[] \phantom{\pcif}$\hData[0]\neq\HashData(\emptyset)$ \pcor
                    \item[] \phantom{\pcif}$\hUnlearn[0]\neq\HashDataUnlearn(\emptyset)\colon$
                    \item \quad \hspace*{0.1cm}$\pcreturn \false$
                    \item \vspace{0.3em} $\pcreturn \true$
				\end{nicodemus}
			\end{minipage}
   			\begin{minipage}[t]{11.8cm}
                \underline{$\circuitTraining(\textbf{public } \hStF[,\cntIteration],\hStF[,\cntIteration-1],\hModel[\cntIteration],\hData[\cntIteration],\hData[\cntIteration-1],\hUnlearn[\cntIteration],\hUnlearn[\cntIteration-1], \textbf{ private } \stF[,\cntIteration-1],\hashsUnlearned[\cntIteration-1],\dataAdd[\cntIteration])~$}
				\begin{nicodemus}
                    \item \codecomment{Check input set of hashed unlearnt data records}
                    \item \pcif $\hUnlearn[\cntIteration-1]\neq\HashDataUnlearn(\hashsUnlearned[\cntIteration-1])\colon$
                    \item \quad \hspace*{0.1cm}$\pcreturn \false$
                    \item \vspace{0.3em}\codecomment{Update and check set of hashed training data records and unlearnt data records}
                    \item $\hashsDataAdd[\cntIteration]\coloneqq \{\HashDataSample(\cntUser,d)\}_{(\cntUser,d)\in\dataAdd[\cntIteration]}$
                    \item $\pcif \hData[\cntIteration]\neq \AppendHashData(\hData[\cntIteration-1],\hashsDataAdd[\cntIteration])$ \pcor $\hUnlearn[\cntIteration]\neq\hUnlearn[\cntIteration-1]$ \pcor $\hashsUnlearned[\cntIteration-1]\cap\hashsDataAdd[\cntIteration]\neq \emptyset\colon$
                    \item \quad \hspace*{0.1cm}$\pcreturn \false$
                    \item \vspace{0.3em} \codecomment{Check input state, perform training and check outputs}
                    \item \pcif $\hStF[,\cntIteration-1]\neq\HashState(\stF[,\cntIteration-1])\colon$
                    \item \quad \pcreturn $\false$
                    \item $(\stF[,\cntIteration],\model[\cntIteration])\coloneqq\trainingFunction(\stF[,\cntIteration-1],\dataAdd[\cntIteration])$
                    \item \pcif $\hStF[,\cntIteration]\neq\HashState(\stF[,\cntIteration])$ \pcor
                    $\hModel[\cntIteration]\neq\HashModel(\model[\cntIteration])\colon$
                    \item \quad \hspace*{0.1cm}$\pcreturn \false$
                    \item \vspace{0.3em} $\pcreturn \true$
				\end{nicodemus}
      \end{minipage}
}}
\end{center}

\section{Completeness Proof}\label{sec:completeness-proof}

\begin{proof}[Proof (of \cref{thm:completeness})] We need to prove the three properties in \cref{def:completeness} capturing the initialization, the proof of training and the proof of unlearning which includes the proof of non-membership.

\vspace{2.5mm}
\noindent\underline{Initialization.} First, running $\Init$ yields the initialized state $\stF[,0]$ and model $\model[0]$ obtained by executing $\initFunction$. Using the hash function to commit to those two values and additionally the empty sets $\hashsData[0]$ and $\hashsUnlearned[0]$, an instance $(\statement_0,\witness_0)\in\relationI$ can be derived and the SNARK proof $\proofI[0]$ can be created using $\SNARKProve$. By correctness of the relation and completeness of the SNARK, $\statement_0$ will be valid for $\com[0]$ and $\SNARKVrfy(\relationI,\ppI,\proofI[0],\statement_0)=1$.

\vspace{2.5mm}
\noindent\underline{Proof of Update.} For the second property, recall the inputs and outputs of $\ProveUpdate$ and $\ProveUnlearn$. The state $\stS[\cntIteration-1]$ contains the state $\stF[,\cntIteration-1]$, the set $\hashsData[\cntIteration-1]$ of hashed data records, the set $\hashsUnlearned[\cntIteration-1]$ of hashed unlearnt data records and the previous commitment $\com[\cntIteration-1]$. In the proof of training, the new state $\stF[,\cntIteration]$ and the new model $\model[\cntIteration]$ are computed by running function $\trainingFunction$ on $\stF[,\cntIteration-1]$ and the set $\dataAdd[\cntIteration]$ of data records to be added. The new sets $\hashsData[\cntIteration]$ and $\hashsUnlearned[\cntIteration]$ are computed from the previous ones and updated with $\dataAdd[\cntIteration]$. The commitment is computed by hashing the four components. 

Then the training instance $(\statement_\cntIteration,\witness_\cntIteration)\in\relationT$ can be derived and the SNARK proof $\proofT[\cntIteration]$ computed. The proof attests (cf.~Appendix~\ref{app:circuits}) that (1) $\model[\cntIteration]$ was computed correctly since $\trainingFunction$ was executed, (2) the training data does not contain unlearnt record since the hashes of the new data records are not contained in $\hashsUnlearned[\cntIteration]$, and (3) the set of unlearnt data records has not changed since the commitments to the unlearnt data records are the same. By correctness of the relation and perfect completeness of the SNARK, we have $\SNARKVrfy(\relationT,\ppT,\proofT[\cntIteration],\statement_\cntIteration)=1$.

Note that there exists a special case where the server is unable to create a proof although the datasets are valid. This is the case whenever there exist two distinct data records $(u,d)\in\dataAdd[\cntIteration]$, $(u',d')\in\dataUnlearn[\cntIteration]$, where $\dataUnlearn[\cntIteration]=\bigcup_{j\in[\cntIteration]}\dataUnlearnAdd[j]$ is the dataset implicitly contained in $\hashsUnlearned[\cntIteration]$, such that $\HashDataSample(u,d)=\HashDataSample(u',d')$. However, we only require computational completeness and assume that the datasets are provided by a PPT adversary. Then this translates to finding a collision for the hash function which happens with negligible probability if the hash function is collision-resistance. Hence, $\VrfyUpdate$ will output $1$ with probability $1-\negl(\lambda)$.

\vspace{2.5mm}
\noindent\underline{Proof of Unlearning.}
Completeness for the proof of unlearning proceeds similar. The new state $\stF[,\cntIteration]$ and the new model $\model[\cntIteration]$ are computed by running function $\unlearningFunction$ on $\stF[,\cntIteration-1]$ and the set $\dataUnlearnAdd[\cntIteration]$ of data records to be deleted. The new sets $\hashsData[\cntIteration]$ and $\hashsUnlearned[\cntIteration]$ are computed from the previous ones and updated by removing and appending $\dataUnlearnAdd[\cntIteration]$, respectively. The commitment is computed by hashing the four components. 

The unlearning instance $(\statement_\cntIteration,\witness_\cntIteration)\in\relationU$ is derived and the SNARK proof $\proofU[\cntIteration]$ that is computed attests (cf.~\cref{fig:circuits-u}) that (1) $\model[\cntIteration]$ was computed correctly since $\unlearningFunction$ was executed, (2) the training data does not contain unlearnt record since we removed the records in $\dataUnlearnAdd[\cntIteration]$ from $\hashsData[\cntIteration]$, and (3) previous set of unlearnt data records is a subset of the updated set since we added the records in $\dataUnlearnAdd[\cntIteration]$ to $\hashsUnlearned[\cntIteration]$ to which we commit. By correctness of the relation and perfect completeness of the SNARK, we have $\SNARKVrfy(\relationU,\ppU,\proofU[\cntIteration],\statement_\cntIteration)=1$ and $\VrfyUnlearn$ will output $1$ with probability $1$.

Finally, consider the algorithm $\ProvePrivUnlearn$. If a data record $(u,d)$ was unlearnt in iteration $\cntIteration$, then its hash is present in $\hashsUnlearned[\cntIteration]$. The proof of non-membership $\proofUnlearn$ consists of the chain path to $(u,d)$ in the chain of $\hashsUnlearned[\cntIteration]$. Let $\com[\cntIteration]$ be the commitment for this iteration, then by correctness of the tree path algorithm, $\VrfyPrivUnlearn$ will output $1$ with probability $1$.
\end{proof}
\section{Security Proof}\label{sec:security-proof}
\nicoresetlinenr
\begin{figure}[t]
	\centering
	\hspace*{-0.2cm}
	\scalebox{0.86}{
		\fbox{\small
			\begin{minipage}[t]{10.65cm}
			    \underline{$\game[0]$-$\game[2]$}
			    \begin{nicodemus}
			        \item $\ppI\gets\SNARKSetup(1^\secpar,\relationI)$
           			\item $\ppT\gets\SNARKSetup(1^\secpar,\relationT)$
			        \item $\ppU\gets\SNARKSetup(1^\secpar,\relationU)$
			        \item $(k,(\cntUser,d),\proofUnlearn,\{\mode[:~~][\cntIteration](\hStF[,\cntIteration],\hModel[\cntIteration],\hData[\cntIteration],\hUnlearn[\cntIteration]),(\statement_\cntIteration,\proofT[\cntIteration])\}_{\cntIteration\in[0:\noIterations]};~$
			        \item[] \hspace*{0.3cm}$ \{\data[\cntIteration]\}_{\cntIteration\in[0:\noIterations]})\gets(\advA\|\extractor)(\relationI,\relationT,\relationU,\ppI,\ppT,\ppU,\initFunction,\trainingFunction,\unlearningFunction,\seed)$
			        \item
           \item \codecomment{Pre-processing}
                    \item  
                    $\dataUnlearn[0]\coloneqq\emptyset$
			        \item $\pcfor \cntIteration\in[\noIterations]$
                    \item \quad \pcif $\mode[][\cntIteration]=\train$
                    \item \qquad $\dataAdd[\cntIteration]\coloneqq\data[\cntIteration]\setminus\data[\cntIteration-1]$
                    \item \quad \pcif $\mode[][\cntIteration]=\unlearn$
                    \item \qquad $\dataUnlearnAdd[\cntIteration]\coloneqq \data[\cntIteration-1] \setminus \data[\cntIteration]$
                    \item \qquad $\dataUnlearn[\cntIteration]\coloneqq \dataUnlearn[\cntIteration-1]\cup\dataUnlearnAdd[\cntIteration]$
                    \item
                    \item \codecomment{Verify commitments}
                    \item $\pcfor i\in[0:\noIterations]\colon$
                    \item \quad $\hashsData[\cntIteration]\coloneqq\{\HashDataSample(\cntUser,d)\}_{(\cntUser,d)\in\data[\cntIteration]}$
                    \item \quad $\hDataP[\cntIteration]\coloneqq\HashData(\hashsData[\cntIteration])$
                    \item \quad \pcif $\hDataP[\cntIteration]\neq\hData[\cntIteration]\colon$
                    \item \qquad \pcreturn $0$
                    \item
                    \item \codecomment{Verify initialization}
                    \item Verify $\statement_0$ valid for $(\hStF[,0],\hModel[0],\hData[0],\hUnlearn[0])$
			        \item \pcif $\pcnot \SNARKVrfy(\relationI,\ppI,\proofI[0],\statement_0)\colon$
                    \item \quad \pcreturn $0$
                    \item
                    \item \codecomment{Re-compute initialization}
                    \item $(\stF[,0],\model[0])\coloneqq\initFunction(\seed)$ \hfill // $\game[1]$-$\game[2]$
                    \item $\pcif \hStF[,0]\neq\HashState(\hStF[,0])$ \pcor $\hModel[0]\neq\HashModel(\model[0])\colon$ \hfill // $\game[1]$-$\game[2]$
                    \item \quad \pcreturn $0$ \hfill // $\game[1]$-$\game[2]$
			    \end{nicodemus}
			\end{minipage}
			\quad 
			\begin{minipage}[t]{9.58cm}
			    \begin{nicodemus}
                    \item \codecomment{Verify proof of training}
                    \item \pcfor $\cntIteration\in[\noIterations]$ s.\,t. $\mode[][\cntIteration]=\train$
                    \item \quad Verify $\statement_{\cntIteration}$ valid for $(\hStF[,\cntIteration],\hModel[\cntIteration],\hData[\cntIteration],\hUnlearn[\cntIteration])$
                    \item \quad \pcif \pcnot $\SNARKVrfy(\relationT,\ppT,\proofT[\cntIteration],\statement_{\cntIteration})\colon$
                    \item \qquad \pcreturn $0$
                    \item
                    \item \codecomment{Verify proof of unlearning}
                    \item \pcfor $\cntIteration\in[\noIterations]$ s.\,t. $\mode[][\cntIteration]=\unlearn$
                    \item \quad Verify $\statement_{\cntIteration}$ valid for $(\hStF[,\cntIteration],\hModel[\cntIteration],\hData[\cntIteration],\hUnlearn[\cntIteration])$
                    \item \quad \pcif \pcnot $\SNARKVrfy(\relationU,\ppU,\proofU[\cntIteration],\statement_{\cntIteration})\colon$
                    \item \qquad \pcreturn $0$
                    \item
                    \item \codecomment{Re-compute state and model and compare to commitment}
                    \item \pcfor $\cntIteration\in[\noIterations]\colon$ \hfill // $\game[1]$-$\game[2]$
                    \item \quad $\pcif \mode[][\cntIteration]=\train$ \hfill // $\game[1]$-$\game[2]$
                    \item \qquad $(\stF[,\cntIteration],\model[\cntIteration])\coloneqq\trainingFunction(\stF[,\cntIteration-1],\dataAdd[\cntIteration])$ \hfill // $\game[1]$-$\game[2]$
                    \item \quad $\pcif \mode[][\cntIteration]=\unlearn$ \hfill // $\game[1]$-$\game[2]$
                    \item \qquad $(\stF[,\cntIteration],\model[\cntIteration])\coloneqq\trainingFunction(\stF[,\cntIteration-1],\dataUnlearnAdd[\cntIteration])$ \hfill // $\game[1]$-$\game[2]$
                    \item \quad $\pcif \hStF[,\cntIteration]\neq\HashState(\stF[,\cntIteration])$ \pcor $\hModel[\cntIteration] \neq\HashModel(\model[\cntIteration])\colon$ \hfill // $\game[1]$-$\game[2]$
			        \item \qquad \pcreturn $0$ \hfill // $\game[1]$-$\game[2]$
                    \item
                    \item \codecomment{Verify proof of non-membership}
                    \item \pcif \pcnot $\VrfyTreePath(\hUnlearn[k],\cntUser,d,\proofUnlearn)\colon$
                    \item \quad \pcreturn $0$
                    \item 
                    \item \codecomment{Check membership of $d$ in $\dataUnlearn[\cntIteration]$}
			        \item \pcfor $\cntIteration\in[k:\noIterations]\colon$ \hfill // $\game[2]$
			        \item \quad \pcif $(\cntUser,d)\notin\dataUnlearn[\cntIteration]\colon$ \hfill // $\game[2]$
			        \item \qquad \pcreturn $0$ \hfill // $\game[2]$
                    \item
                    \item \codecomment{Adversary wins if point unlearned \& re-added}
			        \item $\pcif k<\noIterations ~\pcand (\cntUser,d)\in \dataUnlearnAdd[k] ~\pcand (\cntUser,d)\in\data[\noIterations]\colon$
			        \item \quad \pcreturn $1$
			        \item \pcreturn $0$
			    \end{nicodemus}
			\end{minipage}
	}}
	\caption{
		\emph{Games $\game[0]$-$\game[2]$ for the proof of \cref{thm:security}.} We prove unlearning security for our instantiated protocol $\protocolF$ in Appendix \ref{app:protocol}, where $\admissibleFunction=(\initFunction,\trainingFunction,\unlearningFunction)$ and hyperparameter $\seed$ are fixed by the participating parties and determine relations $\relationI$, $\relationT$ and $\relationU$.} 
		\label{fig:games-proof}
\end{figure}

\begin{proof}[Proof (of \cref{thm:security})]
Let $\advA$ be an adversary against unlearning security (as defined in \cref{fig:sec-unlearn}) of our instantiation. We will first argue that for all $\advA$ there exists an extractor $\extractor$ that outputs the underlying datasets $\data[\cntIteration]$. This follows directly from the knowledge soundness of the SNARK for relations $\relationI$, $\relationT$ and $\relationU$. For this, look at the private inputs to the circuits in \cref{fig:circuits-u} and Appenix~\ref{app:circuits} which translate to the witness. Initialization gives us that $\data[0]=\emptyset$. The proof of training inputs $\dataAdd[\cntIteration]$ and the proof of unlearning inputs $\dataUnlearnAdd[\cntIteration]$ such that we can extract $\data[\cntIteration]=\data[\cntIteration-1]\cup\dataAdd[\cntIteration]$ if $\mode[][\cntIteration]=\train[]$  and $\data[\cntIteration]=\data[\cntIteration-1]\setminus\dataUnlearnAdd[\cntIteration]$ if $\mode[][\cntIteration]=\unlearn[]$.

\medskip
\noindent We continue with the sequence of games given in \cref{fig:games-proof}.

\vspace{2.5mm}
\noindent\underline{Game $\game[0]$.} Let $\game[0]$ be the original game $\SecurityUnlearn$ and $\extractor$ be the extractor. Recall that the adversary must output a sequence of tuples $(k,(u,d),\proofUnlearn,\allowbreak\{\mode[:~~][\cntIteration]\com[\cntIteration],\proofModel[\cntIteration]\}_{\cntIteration\in[0:\noIterations]})$ for some $\noIterations\in\mathbb{N}$, where $\com[\cntIteration]=(\hStF[,\cntIteration],\hModel[\cntIteration],\hData[\cntIteration],\hUnlearn[\cntIteration])$ and $\proofModel[\cntIteration]=(\statement_{\cntIteration},\proofT[\cntIteration])$ for $\cntIteration\in[0:\noIterations]$. We iterate over the winning conditions and return $0$ as soon as one of them is violated (cf.\,\cref{fig:games-proof}). For book-keeping we also compute all sets of unlearnt data points $\dataUnlearn[\cntIteration]$, as well as the sets $\dataAdd[\cntIteration]$, $\dataUnlearnAdd[\cntIteration]$ from $\data[\cntIteration]$ as described for the extractor. Note that this is only a conceptual change at this point and we have
$$\Pr[\game[0]\Rightarrow1]=\Pr[\SecurityUnlearn_{\advA,\extractor,\protocolF,\alldata}(1^\secpar)\Rightarrow1] \ .$$

\vspace{2.5mm}
\noindent\underline{Game $\game[1]$.} In $\game[1]$, we compute the state $\stF[,\cntIteration]$ and the model $\model[\cntIteration]$ for each iteration from the corresponding datasets by applying $\initFunction$, $\trainingFunction$ and $\unlearningFunction$. We then check whether the hashes of state and model correspond to $\hStF[,\cntIteration]$ and $\hModel[\cntIteration]$ in the commitment. If this is not the case, the game outputs $0$. We claim
$$\left|\Pr[\game[1]\Rightarrow1]-\Pr[\game[0]\Rightarrow1]\right|\leq \negl(\secpar) \ .$$
To prove the claim we argue in the following steps:
\begin{itemize}
    \item First, $\proofT[\cntIteration]$ proves that the adversary knows a state $\stF[,\cntIteration]'$ and a dataset ${\dataAddP[\cntIteration]}$ for each proof of training (or a dataset ${\dataUnlearnAddP[\cntIteration]}$ for each proof of unlearning) such that  model $\model[\cntIteration]'$ was computed by applying  function $\trainingFunction$ (or function $\unlearningFunction$) to state $\stF[,\cntIteration-1]'$ and dataset ${\dataAddP[\cntIteration]}$ (or dataset ${\dataUnlearnAddP[\cntIteration]}$). It also proves that the commitment aligns with the inputs. Since the functions are deterministic, we thus have $\hStF[,\cntIteration]=\HashState(\stF[,\cntIteration]')$ and $\hModel[\cntIteration]=\HashData(\model[\cntIteration]')$ as well as $\hData[\cntIteration]=\HashData(\hashsDataP[\cntIteration])$, where $\hashsDataP[\cntIteration]$ is the set of all hashed data records in $\data[\cntIteration]'$.
    
    By soundness of the SNARK, the adversary can only forge a proof for an invalid statement with negligible probability, so we can assume the proof was generated honestly with a witness. By knowledge soundness, the extractor is able to compute this witness such that $\data[\cntIteration]'=\data[\cntIteration]$.
    \item Second, we claim that then $\stF[,\cntIteration]'=\stF[,\cntIteration]$ and $\model[\cntIteration]'=\model[\cntIteration]$ are the actual state and model used for the next iteration. This is true unless the adversary finds a collision in the hash function such that $\HashState(\stF[,\cntIteration]')=\HashState(\stF[,\cntIteration])=\hStF[,\cntIteration]$ or $\HashModel(\model[\cntIteration]')=\HashModel(\model[\cntIteration])=\hModel[\cntIteration]$, which we assume to happen only with negligible probability.
\end{itemize}

\vspace{2.5mm}
\noindent\underline{Game $\game[2]$.} In $\game[2]$, we check whether the data record $(u,d)$ output by $\advA$ is contained in the underlying datasets $\dataUnlearn[\cntIteration]$ of the $k$-th and all subsequent iterations.
We will show that
$$\left|\Pr[\game[2]\Rightarrow1]-\Pr[\game[1]\Rightarrow1]\right|\leq \negl(\secpar) \ .$$
For this we first look again at the SNARK proof $\proofU[\cntIteration]$ and the underlying circuits. If a proof of training is performed, the adversary must prove that $\hUnlearn[\cntIteration]=\hUnlearn[\cntIteration-1]$. This implies---assuming no hash collision occurs---that $\hashsUnlearned[\cntIteration]=\hashsUnlearned[\cntIteration-1]$ and $\dataUnlearn[\cntIteration]=\dataUnlearn[\cntIteration-1]$. If a proof of unlearning is performed, the SNARK proof ensures that $\hashsUnlearned[\cntIteration-1]\subset\hashsUnlearned[\cntIteration]$ and thus $\dataUnlearn[\cntIteration-1]\subset\dataUnlearn[\cntIteration]$, again using collision-resistance of the hash function. Thus, if $(u,d)\in\dataUnlearn[k]$, it must also be true that $(u,d)\in\dataUnlearn[k+1]$, ..., $(u,d)\in\dataUnlearn[\noIterations]$. By soundness of the SNARK, the adversary cannot prove a false statement, so the above claims must hold.

We also know that $(u,d)\in\dataUnlearn[k]$ since the proof of non-membership consists of the path from the hashed data record $(u,d)$ to the hash $\hUnlearn[k]$ contained in the $k$-th commitment. Since the adversary can only win if the proof verifies successfully, we know that in this case the hash value of $(u,d)$, in the following denoted by $h_{u,d}\coloneqq\HashDataSample(u,d)$, must be a node in the hash chain constructed from $\hashsUnlearned[k]$. Unless the adversary finds another data record $(u',d')$ such that $\HashDataSample(u',d')$ maps to the same hash value $h_{u,d}$---which happens with negligible probability---the record $(u,d)$ must be contained in $\dataUnlearn[k]$.\\

Finally, we show that $\Pr[\game[2]\Rightarrow1]\leq\negl(\secpar)$.
For this, recall that $\proofT[\cntIteration]$ also attests that no unlearnt data point is contained in the dataset, in particular that the intersection $\hashsData[\cntIteration]\cap\hashsUnlearned[\cntIteration]$ is empty. Together with the fact that the commitments $\hData[\cntIteration]$ and $\hUnlearn[\cntIteration]$ are constructed from $\hashsData[\cntIteration]$ and $\hashsUnlearned[\cntIteration]$ (due to soundness of the SNARK), the hashed datasets must have been obtained from the corresponding dataset $\data[\cntIteration]$ and $\dataUnlearn[\cntIteration]$ (unless the adversary has found a collision in the hash function). Combining with previous results, this implies that $\data[\cntIteration]\cap\dataUnlearn[\cntIteration]=\emptyset$ for all $i\in[\noIterations]$. As shown above, we know that $(u,d)\in\dataUnlearn[\noIterations]$. The final winning condition requires that $(u,d)\in\data[\noIterations]$. This cannot be the case since it would contradict the fact that the intersection of the two sets is empty, which proves the final claim.

Collecting the probabilities yields
$$\Pr[\SecurityUnlearn_{\advA,\extractor,\protocolF,\alldata}(1^\secpar)]\leq\negl(\secpar) \ ,$$
which concludes the proof of \cref{thm:security}.
\end{proof}
\section{Scalability to Benchmark Datasets}

\label{app:scalability-datasets}
We compute the proof of training for different datasets from the PMLB benchmark suite~\cite{romano-21-pmlb}. Size refers to \#data points $\times$ \#features and {R1CS} refers to \#constraints.\bigskip

\begin{center}
\resizebox{0.6\columnwidth}{!}{
\begin{tabular}{@{}lccccccccc@{}}
\toprule
    \multicolumn{1}{@{}l}{ {{\bf Dataset}} }
&& \multicolumn{2}{@{}c}{ {{\bf Size}} } 
&& \multicolumn{1}{@{}c}{ {{\bf R1CS}} } 
&& \multicolumn{1}{@{}c}{ $\boldsymbol{\SNARKProve}$ }
&& \multicolumn{1}{@{}c}{ $\boldsymbol{\SNARKVrfy}$ }  \\
\midrule
\midrule
Creditscore && $100 $&$ 6$ && 3,986,308 &&  2m 22s &&  0m 47s \\
\rule{0pt}{2ex}%
Patient &&  $88 $&$ 8$ && 4,579,718 &&  2m 28s &&  0m 53s \\
\rule{0pt}{2ex}%
Cy Young &&  $92 $&$ 10$ && 5,903,988 &&  3m 16s &&  1m  9s \\
\rule{0pt}{2ex}%
Corral && $160 $&$ 6$ && 6,347,236 &&  3m 43s &&  1m 15s \\
\rule{0pt}{2ex}%
Lawsuit && $264 $&$ 4$  && 7,190,981 &&  4m  7s &&  1m 27s \\
\rule{0pt}{2ex}%
Breast cancer && $286 $&$ 9$  && 16,514,048 &&  9m 25s &&  3m 18s \\
\rule{0pt}{2ex}%
Monk3 && $554 $&$ 6$  && 21,841,281 && 13m 36s &&  4m 32s \\
\bottomrule
\end{tabular}}

\end{center}

\end{document}